\documentclass[journal]{IEEEtran}
\usepackage{cite}
\usepackage{graphicx}
\usepackage{amsmath}
\usepackage{array}
\usepackage{amssymb}
\usepackage{booktabs}
\usepackage{bbding}
\usepackage{subfigure}
\usepackage{textcomp}
\usepackage{fixltx2e}
\usepackage{url}
\usepackage{color}

\def\ie{{\em i.e.}}
\def\eg{{\em e.g.}}
\def\etal{{\em et al.}}

\begin{document}
\title{WiderPerson: A Diverse Dataset for Dense Pedestrian Detection in the Wild}

\author{Shifeng~Zhang,
        Yiliang~Xie,
        Jun~Wan,~\IEEEmembership{Member,~IEEE,}
        Hansheng~Xia,
        and~Stan~Z.~Li,~\IEEEmembership{Fellow,~IEEE},
        and~Guodong~Guo,~\IEEEmembership{Senior,~IEEE}
\thanks{Shifeng Zhang, Jun Wan and Stan Z. Li are with the Center for Biometric Security Research (CBSR), National Laboratory of Pattern Recognition (NLPR), Institute of Automation Chinese Academy of Sciences (CASIA) and University of Chinese Academy of Sciences (UCAS), Beijing, China. Stan Z. Li is also with the Macau University of Science and Technology, Macau, China (e-mail: \{shifeng.zhang, jun.wan, szli\}@nlpr.ia.ac.cn).}
\thanks{Yiliang Xie is with the University of Southern California (USC), US (e-mail: microos316@gmail.com).}
\thanks{Hansheng Xia is with the College of Energy and Power Engineering, Nanjing University of Aeronautics and Astronautics (NUAA), Nanjing, China (e-mail: hanson\_cha@163.com).}
\thanks{Guodong Guo is with the Institute of Deep Learning, Baidu Research and National Engineering Laboratory for Deep Learning Technology and Application (e-mail: guoguodong01@baidu.com).}
}

\markboth{Journal of IEEE Transactions on MultiMedia}%
{Zhang \MakeLowercase{\textit{et al.}}: WiderPerson: A Diverse Dataset for Dense Pedestrian Detection in the Wild}

\maketitle

\begin{abstract}
Pedestrian detection has achieved significant progress with the availability of existing benchmark datasets. However, there is a gap in the \emph{diversity} and \emph{density} between real world requirements and current pedestrian detection benchmarks: 1) most of existing datasets are taken from a vehicle driving through the regular traffic scenario, usually leading to insufficient diversity; 2) crowd scenarios with highly occluded pedestrians are still under represented, resulting in low density. To narrow this gap and facilitate future pedestrian detection research, we introduce a large and diverse dataset named WiderPerson for dense pedestrian detection in the wild. This dataset involves five types of annotations in a wide range of scenarios, no longer limited to the traffic scenario. There are a total of $13,382$ images with $399,786$ annotations, \ie, $29.87$ annotations per image, which means this dataset contains dense pedestrians with various kinds of occlusions. Hence, pedestrians in the proposed dataset are extremely challenging due to large variations in the scenario and occlusion, which is suitable to evaluate pedestrian detectors in the wild. We introduce an improved Faster R-CNN and the vanilla RetinaNet to serve as baselines for the new pedestrian detection benchmark. Several experiments are conducted on previous datasets including Caltech-USA and CityPersons to analyze the generalization capabilities of the proposed dataset and we achieve state-of-the-art performances on these previous datasets without bells and whistles. Finally, we analyze common failure cases and find the classification ability of pedestrian detector needs to be improved to reduce false alarm and miss detection rates. The proposed dataset is available at \url{http://www.cbsr.ia.ac.cn/users/sfzhang/WiderPerson}.
\end{abstract}


\IEEEpeerreviewmaketitle

\begin{figure}
\begin{centering}
\includegraphics[width=0.5\textwidth]{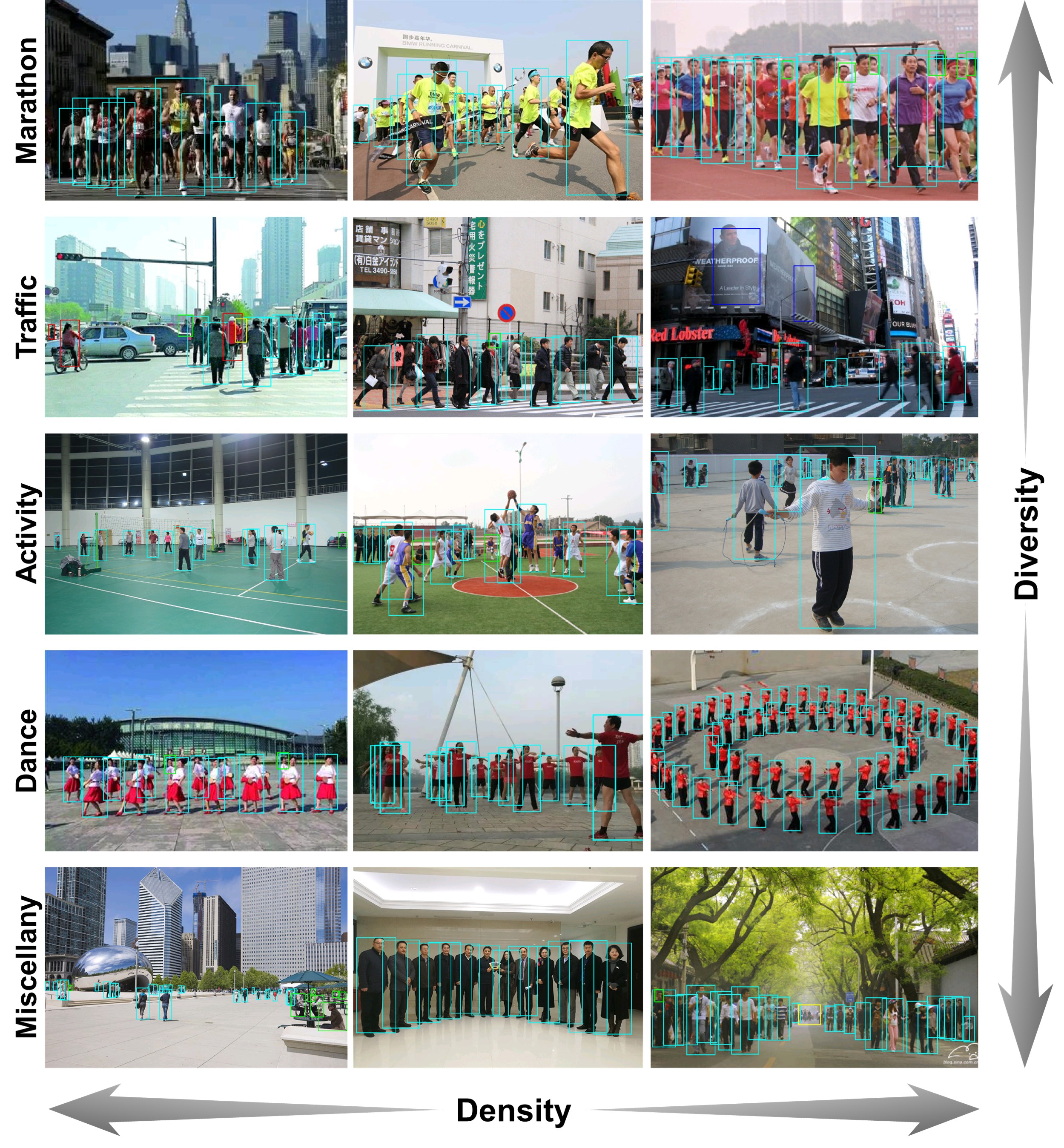}
\par\end{centering}
\caption{The diversity and density of the newly introduced WiderPerson dataset. It can bridge the gap between real world requirements and pedestrian detection benchmarks. For visualization, we use bounding boxes of different colors for pedestrians (Cyan), riders (Red), partially-visible persons (Green), crowd (Yellow) and ignore regions (Blue).}
\label{fig:intro}
\end{figure}

\section{Introduction}
\IEEEPARstart{P}{edestrian} detection is a long-standing problem in computer vision and pattern recognition with extensive applications including security and surveillance, mobile robotics, autonomous driving, and crowd sourcing, to name a few. The accuracy of pedestrian detection systems has a direct impact on these tasks, hence the success of pedestrian detection is of crucial importance. Given an arbitrary image, the goal of pedestrian detection is to determine whether or not there are any pedestrians in the image, and if present, return the image location and extent of each pedestrian. While this appears as an effortless task for human, it is a very difficult task for computers. The challenges associated with pedestrian detection can be attributed to variations in pose, scale and occlusion, which need to be addressed while building pedestrian detection algorithms.

With the remarkable progress over the past few decades, pedestrian detection has been successfully applied in some practical application systems under restricted scenarios. The success of these systems can be attributed to two key steps: (1) advancements in the field of deep Convolutional Neural Network (CNN) which has had a direct impact on many computer vision tasks including pedestrian detection; (2) dataset collection efforts led by different researchers in the community. Furthermore, improvements in detection algorithms have almost always been followed by the publication of more challenging datasets and vice versa. Such synchronous advancement in both steps has led to an even more rapid progress in the field. In terms of pedestrian detection, publicly available benchmark datasets such as Caltech-USA~\cite{DBLP:journals/pami/DollarWSP12}, KITTI~\cite{DBLP:conf/cvpr/GeigerLU12} and CityPersons~\cite{DBLP:conf/cvpr/ZhangBS17} have contributed to spurring interest and progress in pedestrian detection research. Coupled with the development and blooming of deep learning, modern pedestrian detectors~\cite{DBLP:conf/cvpr/ZhangYS18, DBLP:journals/corr/abs-1711-07752, wang2018pedestrian, DBLP:journals/tmm/LiLSXFY18, zhang2018occlusion} have achieved remarkable performance.

Although performance has been significantly improved, it's still difficult to assess for real world, since compared with crowd-counting datasets~\cite{DBLP:conf/cvpr/IdreesSSS13,DBLP:conf/cvpr/ShaoKLW15} designed in the crowded condition, there is a gap in the diversity and density between current existing pedestrian detection benchmarks and real world requirements. On the one hand, most of existing datasets are collected via a vehicle-mounted camera through the regular traffic scenario. This fixed scenario significantly reduces the richness of the foreground and background, leading to low diversity. Specifically, only pedestrians and backgrounds on the road are taken into consideration while the other scenarios are severely under represented. Thus, diversity in pedestrian and background appearances is limited. On the other hand, crowd scenarios with highly occluded pedestrians are still under-represented. As shown in Table~\ref{tab:dataset_statistics}, the Caltech-USA and KITTI datasets have less than one person per image, while the CityPersons dataset has $\sim\negmedspace 7$ persons per image. Even worse, protocols of these datasets allow annotators to ignore the regions with a large number of persons, since exhaustively annotating crowd regions is incredibly difficult and time consuming, resulting in low density and insufficient occlusions cases. To sum up, current pedestrian detection datasets typically contain a few thousand pedestrians with limited variations in diversity and density. These limitations have partially contributed to the failure of some algorithms in coping with heavy occlusion and atypical scenario. Therefore, more challenging datasets similar to the real world are needed to trigger progress and inspire novel ideas.

\begin{table}[t]
\caption{Comparison of pedestrian detection datasets ({\tt Training} subset only). `-' means this term is unlimited and can not be counted.}
\centering
\begin{tabular}{l|cccc}
\toprule[2pt]
 & \scriptsize{Caltech-USA} & \scriptsize{KITTI} & \scriptsize{CityPersons} & \scriptsize{WiderPerson} \\
\midrule[1pt]
\# country & $1$ & $1$ & $3$ & $\bf{-}$ \\
\# city & $1$ & $1$ & $18$ & $\bf{-}$ \\
\# season & $1$ & $1$ & $3$ & $\bf{4}$ \\
\# images & $\bf{42,782}$ & $3,712$ & $2,975$ & $8,000$ \\
\# persons & $13,674$ & $2,322$ & $19,654$ & $\bf{236,073}$ \\
\# ignore regions & $\bf{50,363}$ & $45$ & $6,768$ & $8,979$ \\
\# person/image & $0.32$ & $0.63$ & $6.61$ & $\bf{29.51}$ \\
\# unique persons & $1,273$ & $< 2,322$ & $19,654$ & $\bf{236,073}$ \\
\bottomrule[2pt]
\end{tabular}
\label{tab:dataset_statistics}
\end{table}

To move forward the field of pedestrian detection, we introduce a diverse and dense pedestrian detection dataset called WiderPerson. It consists of $13,382$ images with $399,786$ annotations, \ie, $29.87$ annotations per image, varying largely in scenario and occlusion, as shown in Fig.~\ref{fig:intro}. Besides, the annotations have five fine-grained labels, \ie, pedestrians, riders, partially-visible persons, crowd, and ignore regions. These high quality annotations provide a rich diverse dataset and enable new experiments both for training better models, and as new test benchmark. We split the proposed WiderPerson dataset into three subsets (training, validation, and testing sets). Annotations of training and validation will be public, and an online benchmark will be set-up. We show an example of using the proposed WiderPerson dataset through proposing an improved Faster R-CNN~\cite{DBLP:journals/pami/RenHG017}, which consists of finer feature map, ignore region and tiny pedestrian handling, Region of Interest (RoI) feature enhancing and dynamic sample strategy to deal with large density and diversity variations. The cross-dataset generalization results of the proposed WiderPerson dataset show that it is an effective training source for pedestrian detection and we achieve state-of-the-art performance on existing Caltech-USA and CityPersons datasets. 

For clarity, the main contributions of this work can be summarized as three-fold:
\begin{itemize}
\item We propose the WiderPerson dataset, which provides a large number of highly diverse and dense bounding box annotations for pedestrian detection. 
\item We build an improved Faster R-CNN to show an example of using WiderPerson, which consists of some improvements to deal with large density and diversity variations.
\item We prove the generalization capabilities of detectors trained with the new dataset and achieve state-of-the-art performance on Caltech-USA and CityPersons datasets.
\end{itemize}

The rest of the paper is organized as follows. Section \ref{2} reviews the related work. Description of the WiderPerson dataset is presented in Section \ref{3}. Section \ref{4} introduces our proposed baseline detector and Section \ref{5} shows the experimental results. Section \ref{6} concludes the paper.

\section{Related Work} \label{2}
\subsection{Dataset}

In the last decade, several datasets have been created for pedestrian detection training and evaluation. The GM-ATCI dataset~\cite{DBLP:conf/ivs/SilbersteinLKG14} is collected using a vehicle-mounted standard automotive rear-view display camera for evaluating rear-view pedestrian detection. The INRIA dataset~\cite{DBLP:conf/cvpr/DalalT05} is one of the most popular static pedestrian detection datasets. The USC dataset~\cite{DBLP:conf/iccv/WuN07} consists of a number of fairly small pedestrian datasets taken largely from surveillance video. The ETH dataset~\cite{DBLP:conf/iccv/EssLG07} is captured from a stereo rig mounted on a stroller in the urban. The CVC-ADAS dataset~\cite{geronimo2007adaptive} contains pedestrian videos acquired on-board, virtual-world pedestrians (with part annotations) and occluded pedestrians. The NICTA dataset~\cite{overett2008new} is a large scale urban dataset collected in multiple cities and countries, it has no motion and tracking information but significant number of unique pedestrians. The Daimler dataset~\cite{DBLP:journals/pami/EnzweilerG09} is captured in an urban setting and has tracking information and a large number of labelled bounding boxes. The TUD-Brussels dataset~\cite{DBLP:conf/cvpr/WojekWS09} contains image pairs recorded in a crowded urban setting with an onboard camera. These datasets represent early efforts to collect pedestrian datasets.

Althought these early datasets have contributed to spurring interest and progress of pedestrian detection, however, as algorithm performance improves, they are replaced by the larger and richer datasets. The Tsinghua-Daimler Cyclist (TDC) dataset~\cite{DBLP:conf/ivs/LiFYXBPLG16} focuses on cyclists recorded from a vehicle-mounted stereo vision camera, containing a large number of cyclists varying widely in appearance, pose, scale, occlusion and viewpoint. In~\cite{DBLP:conf/cvpr/HwangPKCK15}, a multi-spectral dataset for pedestrian detection is introduced, combining RGB and infrared modalities. The Caltech-USA~\cite{DBLP:journals/pami/DollarWSP12} dataset consists of approximately $10$ hours of $640\times480$ $30$Hz video taken from a vehicle driving through regular traffic in an urban environment, which has been extended by~\cite{DBLP:journals/pami/ZhangBOHS18} with corrected annotations. The KITTI~\cite{DBLP:conf/cvpr/GeigerLU12} dataset focuses autonomous driving and is collected via a standard station wagon with two high-resolution color and grayscale video cameras, around the mid-size city of Karlsruhe, in rural areas and on highways, up to $15$ cars and $30$ pedestrians are visible per image. The CityPersons~\cite{DBLP:conf/cvpr/ZhangBS17} dataset is recorded by a car traversing $27$ different cities and provides high quality bounding boxes with larger portions of occluded persons. The EuroCity Persons dataset~\cite{DBLP:journals/corr/abs-1805-07193} provides a large number of highly diverse, accurate and detailed annotations of pedestrians, cyclists and other riders in $31$ cities of $12$ European countries.

Despite the prevalence of these datasets, they all suffer a problem of from low diversity. Most of existing datasets are collected via a vehicle-mounted camera through the regular traffic scenario. The diversity in pedestrian and background appearances is limited. Another weakness of both datasets is that the crowd scenarios are significantly under represented, resulting in insufficient occlusions cases. The paper aims at solving these two problems via proposing a diverse and dense pedestrian detection dataset, which can narrow the gap in the diversity and density between real world requirements and current pedestrian detection benchmarks to better evaluate detectors in the wild. Besides, the proposed dataset is also very useful for training a re-detector for dealing with tracking loss for pedestrian tracking~\cite{DBLP:journals/tip/SundaresanC09, DBLP:journals/tip/LanZYC18}.

\subsection{Method}
{\flushleft \textbf{Generic Object Detection.}}
Early generic object detection methods rely on the sliding window paradigm based on the hand-crafted features and classifiers to find the objects of interest. In recent years, with the advent of deep Convolutional Neural Network (CNN), a new generation of more effective object detection methods based on CNN significantly improve the state-of-the-art performances, which can be roughly divided into two categories, \ie, the one-stage approach and the two-stage approach. The one-stage approach~\cite{DBLP:conf/eccv/LiuAESRFB16,DBLP:conf/cvpr/ZhangSF18} directly predicts object class label and regresses object bounding box based on the pre-tiled anchor boxes using deep CNN. The main advantage of the one-stage approach is its high computational efficiency. In contrast to the one-stage approach, the two-stage approach~\cite{DBLP:journals/pami/RenHG017, DBLP:journals/tmm/LiWLDXFY17} always achieves top accuracy on several benchmarks, which first generates a pool of object proposals by a separated proposal generator, and then predicts the class label and accurate location and size of each proposal.

{\flushleft \textbf{Pedestrian Detection.}}
Even as one of the long-standing problems in computer vision field with an extensive literature, pedestrian detection still receives considerable interests with a wide range of applications. A common paradigm~\cite{DBLP:conf/bmvc/DollarTPB09,DBLP:conf/cvpr/YanZLLL13,DBLP:conf/cvpr/ZhangBC14} to deal with this problem is to train a pedestrian detector that exhaustively operates on the sub-images across all locations and scales. Dalal and Triggs~\cite{DBLP:conf/cvpr/DalalT05} design the Histograms of Oriented Gradient (HOG) descriptors and Support Vector Machine (SVM) classifier for human detection. Doll{\'{a}}r \etal~\cite{DBLP:journals/pami/DollarABP14} demonstrate that using features from multiple channels can greatly improve the performance. Zhang \etal~\cite{DBLP:conf/cvpr/ZhangBS15} provide a systematic analysis for the filtered channel features, and find that with the proper filter bank, filtered channel features can reach top detection quality. Paisitkriangkrai \etal~\cite{DBLP:conf/eccv/PaisitkriangkraiSH14} design a new feature built on low-level features and spatial pooling, and directly optimize the partial area under the Receiver Operating Characteristic (ROC) curve for better performance.

Recently, CNN-based detectors~\cite{DBLP:conf/cvpr/SermanetKCL13,DBLP:conf/cvpr/HosangOBS15,DBLP:conf/cvpr/TianLWT15,DBLP:conf/iccv/BrazilYL17} have become a predominating trend in the field of pedestrian detection. Sermanet \etal~\cite{DBLP:conf/cvpr/SermanetKCL13} present an unsupervised method using the convolutional sparse coding to pre-train CNN for pedestrian detection. In~\cite{DBLP:conf/iccv/CaiSV15}, a complexity-aware cascaded detector is proposed for an optimal trade-off between accuracy and speed. Angelova \etal~\cite{DBLP:conf/bmvc/AngelovaKVOF15} combine the ideas of fast cascade and a deep network to detect pedestrian. Yang \etal~\cite{DBLP:conf/cvpr/YangCL16} use scale-dependent pooling and layer-wise cascaded rejection classifiers to detect objects efficiently. Zhang \etal~\cite{DBLP:conf/eccv/ZhangLLH16} present an effective pipeline for pedestrian detection via extracting self-learned features from the Region Proposal Network (RPN)~\cite{DBLP:journals/pami/RenHG017} followed by a boosted decision forest. Cai \etal~\cite{DBLP:conf/eccv/CaiFFV16} propose an architecture which uses different levels of features to detect persons at various scales. Mao \etal~\cite{DBLP:conf/cvpr/MaoXJC17} present a multi-task network architecture to jointly learn pedestrian detection with the given extra features. Li \etal~\cite{DBLP:journals/tmm/LiLSXFY18} use multiple built-in sub-networks to adaptively detect pedestrians across scales. Brazil \etal~\cite{DBLP:conf/iccv/BrazilYL17} exploit weakly annotated bounding boxes via a segmentation infusion network to achieve considerable performance gains.

Occlusion is one of the most significant challenges in compute vision, especially for pedestrian detection, which increases the difficulty in pedestrian localization. Several methods~\cite{DBLP:conf/cvpr/OuyangW12,DBLP:conf/iccv/TianLWT15,DBLP:conf/accv/ZhouY16,DBLP:conf/iccv/WuN05,DBLP:conf/eccv/DuanAL10} use part-based model to describe the pedestrian in occlusion handling, which learn a series of part detectors and design some mechanisms to fuse the part detection results to localize partially occluded pedestrians. Besides the part-based model, Leibe \etal~\cite{DBLP:conf/cvpr/LeibeSS05} propose an implicit shape model to generate a set of pedestrian hypotheses that are further refined to obtain the visible regions. Wang \etal~\cite{DBLP:conf/iccv/WangHY09} divide the template of pedestrian into a set of blocks and conduct occlusion reasoning by estimating the visibility status of each block. Ouyang \etal~\cite{DBLP:conf/cvpr/OuyangW13} exploit multi-pedestrian detectors to aid single-pedestrian detectors to handle partial occlusions, especially when the pedestrians gather together and occlude each other in real-world scenarios. In~\cite{DBLP:conf/cvpr/PepikSGS13}, a set of occlusion patterns of pedestrians are discovered to learn a mixture of occlusion-specific detectors. Zhou \etal~\cite{DBLP:conf/iccv/ZhouY17} propose to jointly learn part detectors to exploit part correlations and reduce the computational cost. Wang \etal~\cite{DBLP:journals/corr/abs-1711-07752} introduce a new bounding box regression loss to detect pedestrians in crowd scenarios.

\section{Proposed WiderPerson Dataset}\label{3}
In this section, we present our WiderPerson dataset from aspects of collection process, annotation tool, annotation method, various statistical information and benchmarking.

\subsection{Data Collection}
For the diversity of our dataset, we crawl images from multiple image search engines ranging from Google, Bing, and Baidu. Combined with specially-designed keywords, one of the prominent advantages of using different image search engines together is the collected images possess diverse features in cities, events, and scenarios. We design more than $50$ keywords (\eg, pedestrian, cyclist, walking, running, marathon, square dance and group photo) during the crawling process and obtain $\sim\negmedspace50,000$ images as our candidate images. To prevent the duplication of images, we leverage a simple but powerful mechanism, the pHash~\cite{DBLP:conf/ccs/MihcakV01}, along with the union find, for the removal of the repetitions. Moreover, images with sparse distribution of people are filtered out to keep the difficulties of our dataset. Finally, we have $13,382$ images remained, and they are randomly split into training, validation and testing subsets with $8,000$, $1,000$ and $4,382$ images, respectively.

\subsection{Annotation Tool}
\begin{figure}[t]
\begin{centering}
\includegraphics[width=0.49\textwidth]{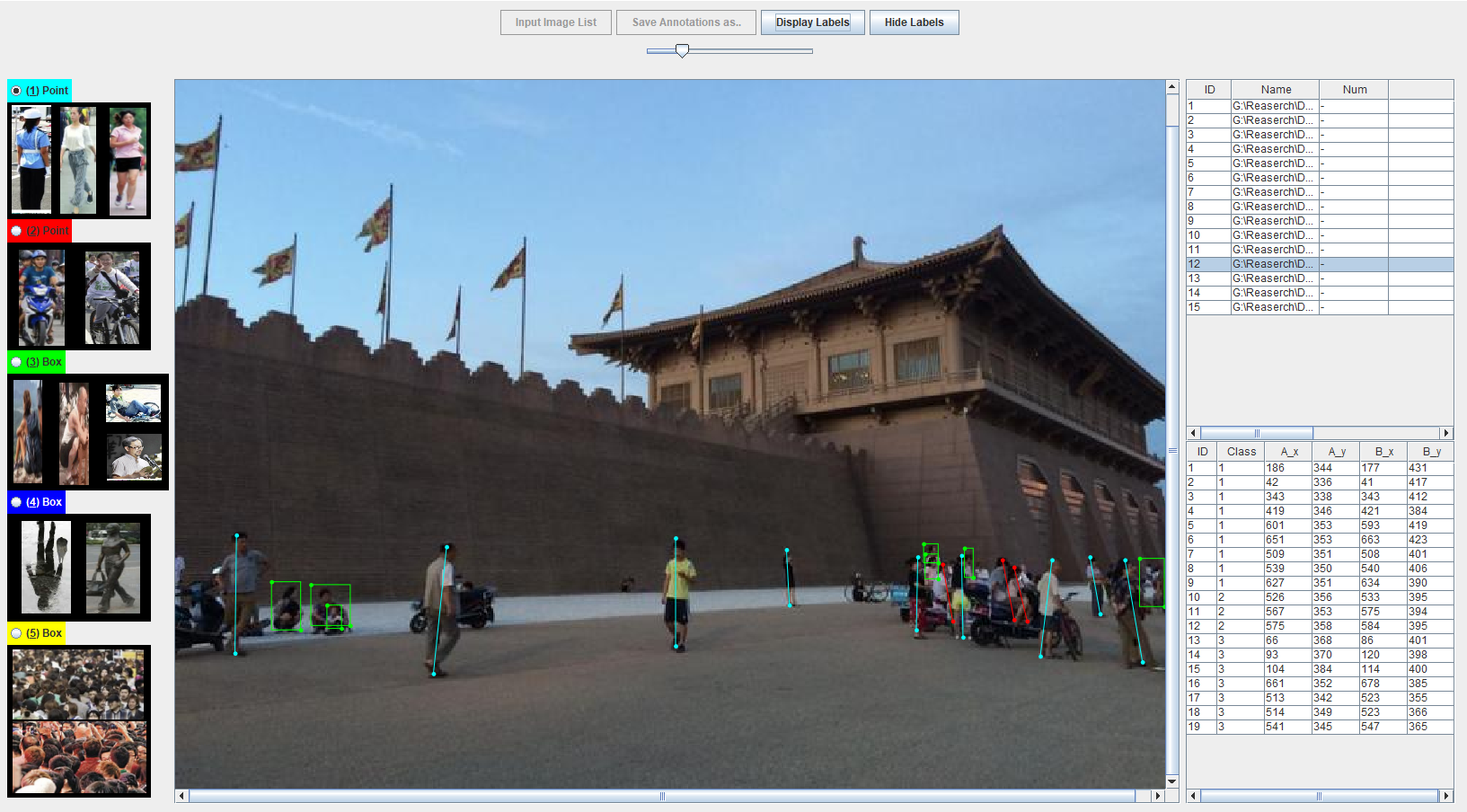}
\par\end{centering}
\caption{Graphical User Interface (GUI) of our annotation tool.}
\label{fig:anno_tool}
\end{figure}

We design a new annotation tool whose Graphical User Interface (GUI) is illustrated in Fig.~\ref{fig:anno_tool}. It is written using JavaScript and built with a very responsive design. The list of images that need to be marked is displayed on the upper right side. For the selected image to be annotated, the tool displays five kinds of annotation examples on the left side to help annotators to mark. These five different types of annotations are labelled with different colors to better distinguish. All the labelled annotations are shown on the lower right side and annotators can select any labelled annotation to display and correct. To complete the annotation, the user shall next adjust the position of the bounding boxes. For this purpose, the keyboard arrows shall be used. More precisely, the left, right, up and down keys should be used in order to shift the annotations on the image. To help annotators, there are two buttons (display and hide labels) at the top used to make labelled annotations optionally visible while annotating. In addition, a zooming feature at the top can be used to zooming-in the corresponding image and annotations. This is implemented in order to make it easier for the annotators to obtain more precise locations of the annotations. After getting used to the annotation process, annotators become more and more precise on these steps, which significantly reduces the time required to annotate as fewer adjustments are required. Besides, to ensure that profit is not prioritized over the accuracy and the precision of the annotations, we are highly involved in the process and all annotators must pass the strict annotation testing.

\begin{figure}[t]
\centering
\subfigure[Image]{
\label{fig:examples-bb-annos-a}
\begin{minipage}[b]{0.31\linewidth}
\includegraphics[width=1\linewidth]{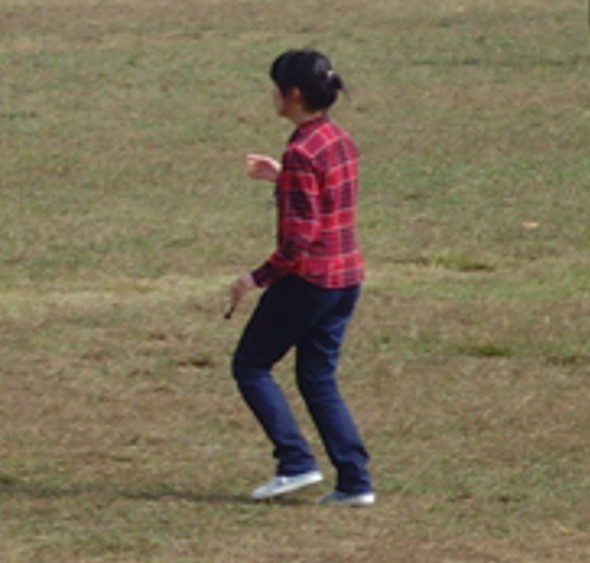}\vspace{5pt}
\includegraphics[width=1\linewidth]{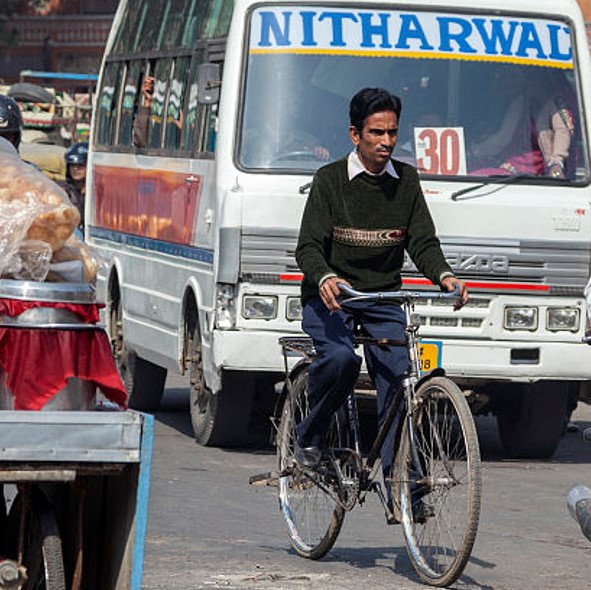}
\end{minipage}}
\subfigure[Labelled Annotation]{
\label{fig:examples-bb-annos-b}
\begin{minipage}[b]{0.31\linewidth}
\includegraphics[width=1\linewidth]{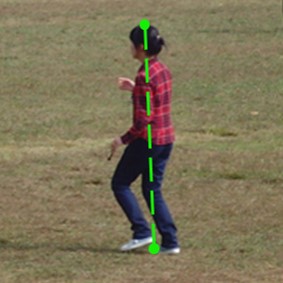}\vspace{5pt}
\includegraphics[width=1\linewidth]{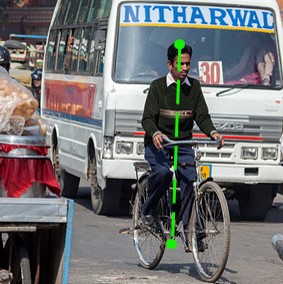}
\end{minipage}}
\subfigure[Generated Bbox]{
\label{fig:examples-bb-annos-c}
\begin{minipage}[b]{0.31\linewidth}
\includegraphics[width=1\linewidth]{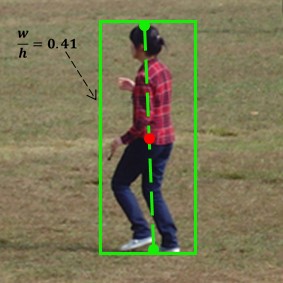}\vspace{5pt}
\includegraphics[width=1\linewidth]{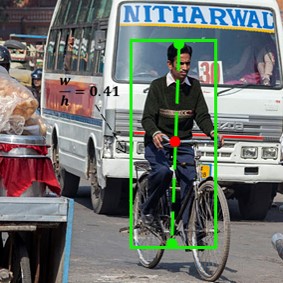}
\end{minipage}}
\caption{Illustration of bounding box annotations for pedestrians and riders. For each target, the top of the head and middle of the feet is drawn by the annotator. An aligned bounding box is automatically generated using the fixed aspect ratio (0.41).}
\label{fig:examples-bb-annos}
\end{figure}

\subsection{Image Annotation}
Our Annotations are finely classified into five categories: pedestrians, riders, partially-visible persons, crowd and ignore regions. The annotation process contains the two steps:

\begin{itemize}
\item[1.] Annotators are asked to thoroughly search across the whole image for individuals, and annotate them for using the similar protocol from~\cite{DBLP:conf/cvpr/ZhangBS17}. For pedestrians and riders (shown in Fig.~\ref{fig:examples-bb-annos-a}), we generate a bounding box by drawing a line across one's head and the middle point between feet, as shown in Fig.~\ref{fig:examples-bb-annos-b}. A bounding box aligned to the center of the line is then generated with an aspect ratio of $0.41$ (defined as w/h), as shown in Fig.~\ref{fig:examples-bb-annos-c}. For partially-visible persons, including individuals that are heavily-occluded or with unusual poses and viewpoints, we mark them using bounding boxes with unconstrained aspect ratios. The crowd in our dataset plays another critical role contributing to the variances and difficulties. Similar to the partially-visible persons, we also annotate a group of people using a tightly bounded rectangle. Finally, we annotate regions containing fake human, for instance, human on the posters, reflections, mannequin and statues, etc.

\item[2.] After the above-mentioned annotating process, to ensure the quality of the labels, we perform three-fold cross-validation to check the annotations strictly. Each image is intuitively marked as either correct or erroneous by three different annotators, and if it marked as erroneous by more than half of the annotators, it would be re-annotated until it passes the check. Fig.~\ref{fig:intro} shows some exemplary final annotations.
\end{itemize}

\subsection{Dataset Statistic} \label{sec:dataset_stat}

\begin{figure}
\begin{centering}
\includegraphics[width=0.43\textwidth]{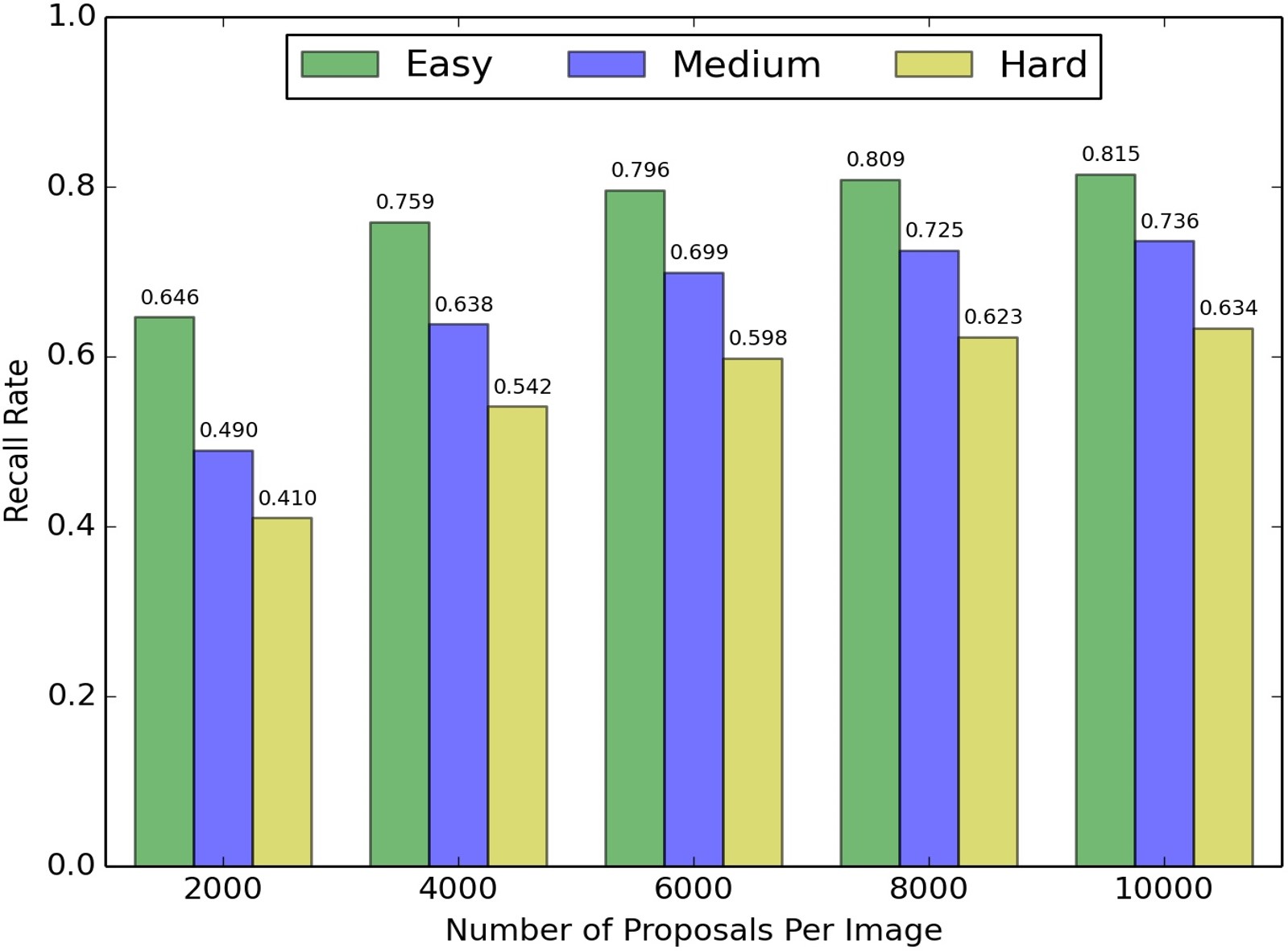}
\par\end{centering}
\caption{Recall rate with different number of proposals. Proposals are generated by using Edgebox~\cite{DBLP:conf/eccv/ZitnickD14}. Lower recall rate implies higher difficulty. We show histograms of detection rate over the number of proposal for different subsets.}
\label{fig:wp_stats}
\end{figure}

{\flushleft \textbf{Capacity. }}
The number of bounding box annotations provided by our WiderPerson dataset is shown in table~\ref{tab:Statistics-bb}, which illustrates the capacity of WiderPerson dataset. In a total of $13,382$ images, there are $\sim\negmedspace 386k$ person and $\sim\negmedspace 13k$ ignore region annotations in the WiderPerson dataset. The number of annotations is more than $10\times$ boosted compared with previous challenging pedestrian detection dataset like CityPersons. The total number of persons is also noticeably larger than the others. We randomly select $8000/1000/4382$ images as training/validation/testing subsets. Following the principle in WIDER FACE~\cite{DBLP:conf/cvpr/YangLLT16}, we define three levels of difficulty: `Easy' ($\geq100$ pixels), `Medium' ($\geq50$ pixels), `Hard' ($\geq20$ pixels) according to the physical height of ground-truth bounding boxes. As shown in the Fig.~\ref{fig:wp_stats}, we utilize EdgeBox~\cite{DBLP:conf/eccv/ZitnickD14} to evaluate their detection rates with different number of proposals. The average recall rates for these three levels are $81.5\%$, $73.6\%$ and $63.4\%$ with $10,000$ proposal per image.

\begin{table}[t]
\caption{Statistics of annotations on WiderPerson dataset.}
\label{tab:Statistics-bb}
\begin{centering}
\setlength{\tabcolsep}{7.5pt}
\begin{tabular}{c|ccc|c}
\toprule[2pt]
 & Training & Validation & Testing & Sum \\
\midrule
{\# images} & $8,000$ & $1,000$ & $4,382$ & $13,382$ \\
\# persons & $236,073$ & $27,762$ & $122,518$ & $386,353$ \\
\# ignore regions & $8,979$ & $661$ & $3,793$ & $13,433$ \\
\# person/images & $29.51$ & $27.76$ & $27.96$ & $28.87$ \\
\bottomrule[2pt]
\end{tabular}
\par\end{centering}
\end{table}

{\flushleft \textbf{Scale. }}
To analyse the scale characteristic across different datasets, we use the probability density function (PDF) to specify the probability of scale falling within a particular range of values, which can specify the distribution of scales. To this end, we group the persons by their image size (height in pixels) into some scale bins. As can be observed from Fig.~\ref{fig:scale}, Caltech-USA and CityPersons have a limited scale distribution, most of their annotations are between $30\negmedspace\sim\negmedspace100$ pixels in height. In contrast, our WiderPerson dataset covers a much wider range of scale and the distribution of persons at all scales is relatively uniform.

\begin{figure}[t]
\centering
\label{fig:scale}
\includegraphics[width=0.45\textwidth]{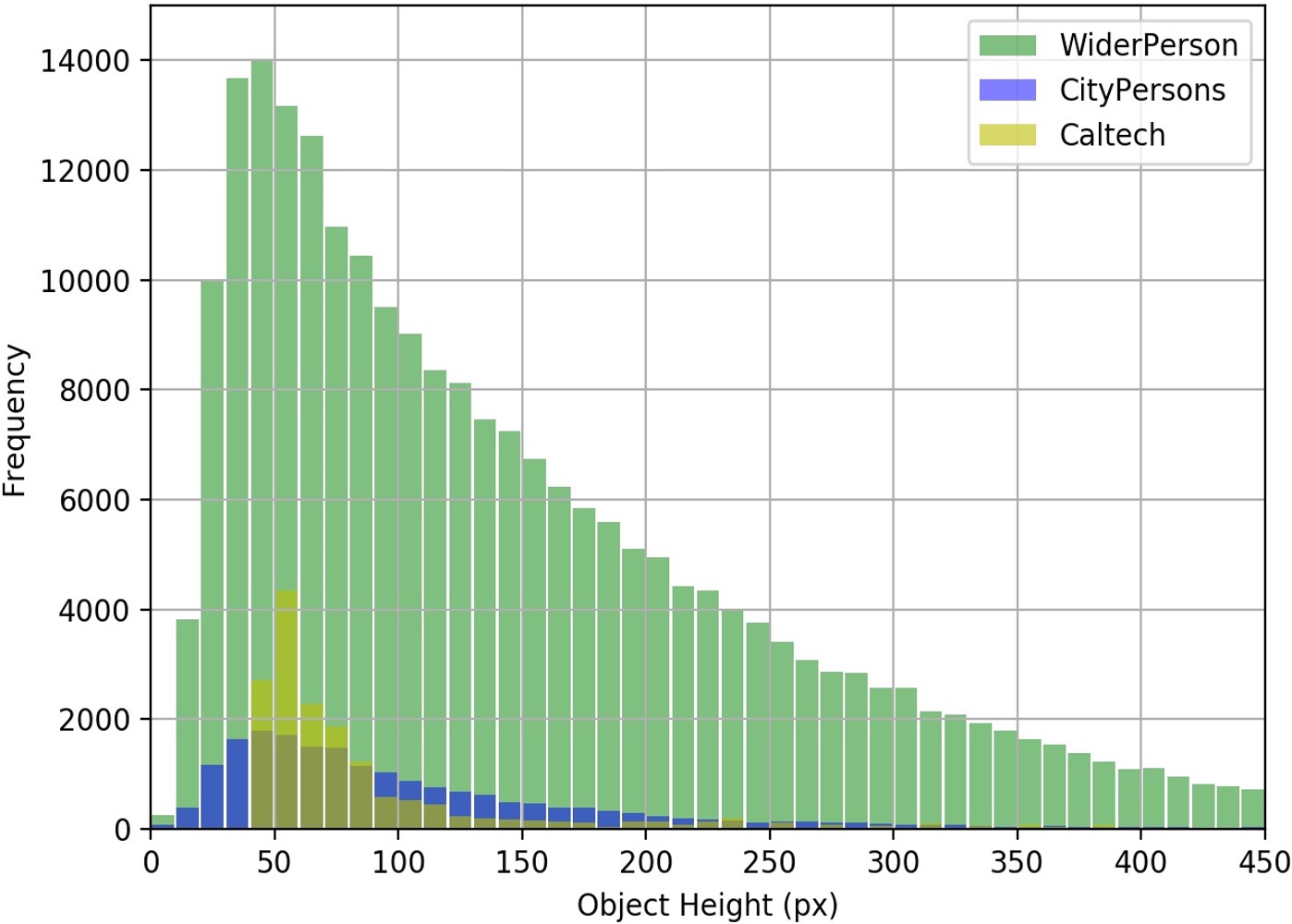}
\caption{Scale distribution of different dataset. We use the probability density function (PDF) to specify the probability of scale falling within a particular range of values.}
\label{fig:scale}
\end{figure}

\begin{figure*}[t]
\centering
\subfigure[Caltech-USA]{
\label{fig:Caltech-USA}
\includegraphics[height=0.19\textwidth]{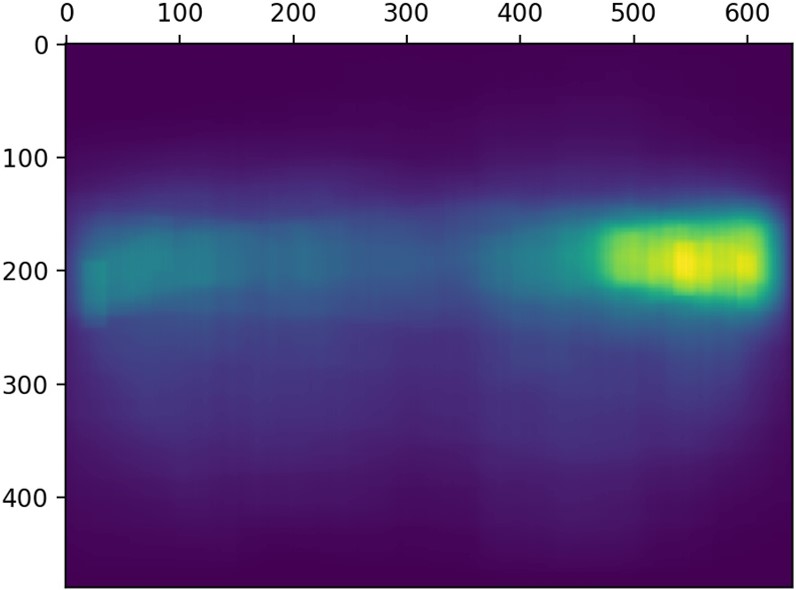}}
\subfigure[CityPersons]{
\label{fig:CityPersons}
\includegraphics[height=0.19\textwidth]{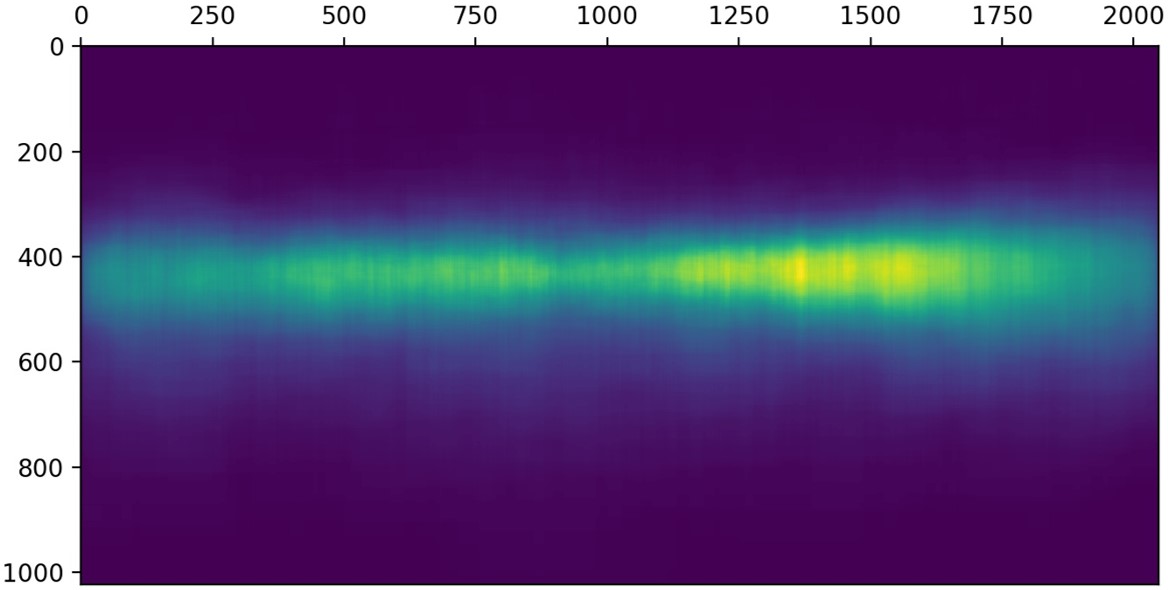}}
\subfigure[WiderPerson]{
\label{fig:WiderPerson}
\includegraphics[height=0.19\textwidth]{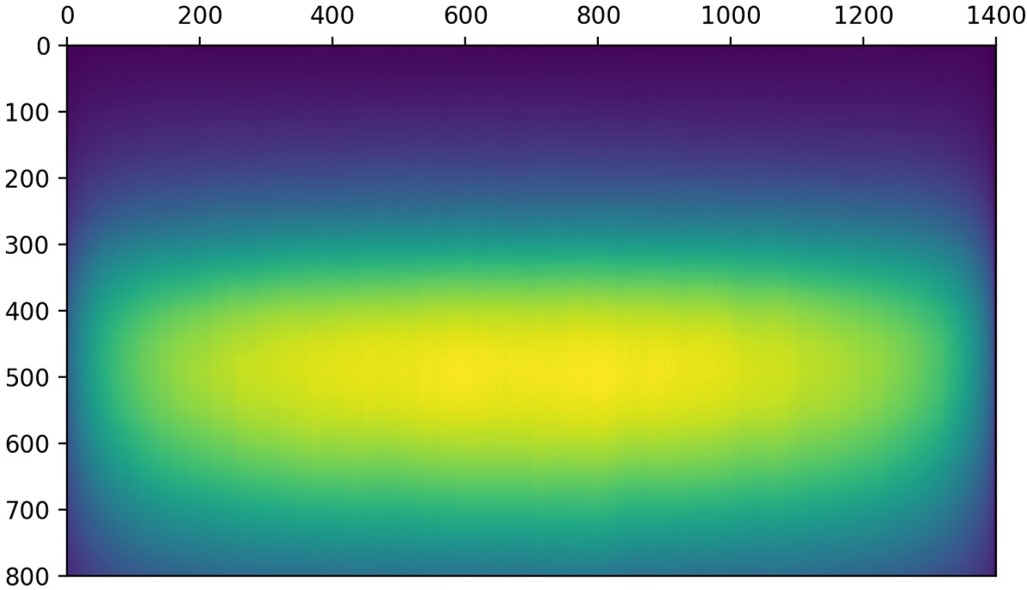}}
\caption{The location distribution of pedestrians on the image. Pedestrians on the Caltech-USA and CityPersons dataset are distributed in a narrow band across the center of the image, while WiderPerson has an uniform location distribution.}
\label{fig:heatmap}
\end{figure*}

{\flushleft \textbf{Density. }}
In terms of density, on average there are $\sim\negmedspace 28.87$ persons per image in WiderPerson dataset, as shown in the fourth line of Table~\ref{tab:Statistics-bb}. We also report the density from the existing datasets in Table~\ref{tab:density}. Obviously, WiderPerson dataset is of much higher crowdness compared with all previous datasets. Caltech-USA suffers from extremely low-density, for that on average there is only $\sim\negmedspace1$ person per image. The number in CityPersons reaches $\sim\negmedspace7$, a significant boost while still not dense enough. Both of them are insufficient to serve as an ideal benchmark for the challenging crowd scenes. Thanks to the pre-filtering and annotation protocol of our dataset, WiderPerson can reach a much better density. As shown in Table.~\ref{tab:Statistics-bb}, we notice the density of persons are consistent across training/validation/testing subsets.

\begin{table}[t]
\caption{Density comparison between widely used pedestrian detection datasets. It demonstrates the number and proportion of images that contain $\ge$ \# persons in different datasets.}
\label{tab:density}
\setlength{\tabcolsep}{7pt}
\begin{center}
\begin{tabular}{c|cc|cc|cc}
\toprule[2pt]
{\# persons} & \multicolumn{2}{c|}{Caltech-USA} & \multicolumn{2}{c|}{CityPersons} & \multicolumn{2}{c}{WiderPerson} \\
\midrule
$\ge$1 & {7,839} & {18.3\%} & {2,482} & {83.4\%} & \textbf{8,000} & {100.0\%} \\
$\ge$2 & {3,257} & {7.6\%} & {2,082} & {70.0\%} & \textbf{7,999} & {100.0\%} \\
$\ge$3 & {1,265} & {3.0\%} & {1,741} & {58.5\%} & \textbf{7,998} & {100.0\%} \\
$\ge$5 & {282} & {0.7\%} & {1,225} & {41.2\%} & \textbf{7,994} & {99.9\%} \\
$\ge$10 & {36} & {0.1\%} & {610} & {20.5\%} & \textbf{7,924} & {99.1\%} \\
$\ge$20 & {0} & {0.0\%} & {227} & {7.6\%} & \textbf{5,145} & {64.3\%} \\
$\ge$30 & {0} & {0.0\%} & {94} & {3.2\%} & \textbf{2,564} & {32.1\%} \\
\bottomrule[2pt]
\end{tabular}
\end{center}
\end{table}

{\flushleft \textbf{Diversity. }}
Diversity is an important factor of a dataset. We compare the diversity of Caltech-USA, CityPersons and WiderPerson in Table.~\ref{tab:dataset_statistics}. Since CityPersons testing set annotations are not publicly available, we only consider the training subset for a fair comparison. The Caltech-USA and KITTI datasets are recorded in one city at one season, and the CityPersons dataset is recorded across $18$ cities, $3$ countries and seasons, while our WiderPerson dataset has no limitations on these conditions. WiderPerson contains person in a wide range of scenarios, while Caltech-USA and CityPersons are all recorded by a car traversing on streets. In order to visualize the diversity of annotations on the different datasets, we count the location distribution of persons, \ie, iterating over all person annotations, for each location, if it is inside one annotation, then its count plus $1$. The images of Caltech and CityPersons have a fixed resolution ($640\times480$ and $2048\times1024$, respectively), while our dataset varies in size, so we resize all images into the same resolution ($1400\times800$) to count the location distribution of persons. Fig.~\ref{fig:heatmap} shows the location distribution of persons for different dataset in the way of heat map. We can see that persons on the Caltech-USA and CityPersons dataset are distributed in a narrow band across the center of the image, \ie, persons are concentrated on two sides of the road and mostly appear at the right side, since their images are collected by a biased data collection method that the car drives under the right-handed traffic condition. In contrast, our WiderPerson dataset has a uniform location distribution and persons appear in any position except the upper part (\ie, the sky).

\begin{figure}
\begin{centering}
\includegraphics[width=0.35\textwidth]{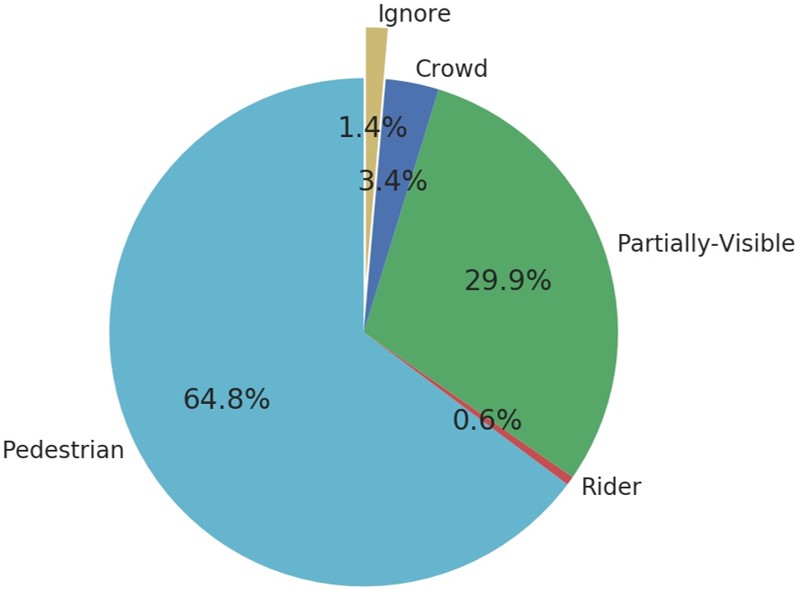}
\par\end{centering}
\caption{Fine-grained person categories on WiderPerson.}
\label{fig:human-categories}
\end{figure}

Also, the number of identical persons is another important evidence of diversity. As reported in the fifth line in Table~\ref{tab:dataset_statistics}, the number of identical persons amounts up to $\sim\negmedspace 236k$ in our WiderPerson dataset. In contrast, the Caltech-USA dataset only contains $\sim\negmedspace 1,300$ unique pedestrians, since images in Caltech-USA are not sparsely sampled, resulting in less amount of identical persons. While CityPersons frames are sampled very sparsely and each person is considered as unique. Like CityPersons, each person on our dataset can be considered as unique, but one more order of magnitude. Besides, WiderPerson also provides fine-grained labels for persons. As shown in Fig.~\ref{fig:human-categories}, pedestrians are the majority ($64.8\%$). Partially-visible persons account for $29.9\%$ since our dataset is dense. Although riders only occupy $0.6\%$, the absolute numbers are still considerable, as we have a large pool of $\sim\negmedspace 236k$ persons.

\begin{table}[t]
\caption{Comparison of pair-wise overlap between two person instances.} \label{tab:PairOverlap}
\begin{center}
\begin{tabular}{c|ccc}
\toprule[2pt]
{\ \ \ pair/image\ \ \ } & {Caltech-USA} & {CityPersons} & {WiderPerson} \\
\midrule
{IoU$>$0.3} & {0.06} & {0.96} & \bf{9.21} \\
{IoU$>$0.4} & {0.03} & {0.58} & \bf{4.78} \\
{IoU$>$0.5} & {0.02} & {0.32} & \bf{2.15} \\
{IoU$>$0.6} & {0.01} & {0.17} & \bf{0.81} \\
{IoU$>$0.7} & {0.00} & {0.08} & \bf{0.24} \\
{IoU$>$0.8} & {0.00} & {0.02} & \bf{0.06} \\
{IoU$>$0.9} & {0.00} & {0.00} & \bf{0.01} \\
\bottomrule[2pt]
\end{tabular}
\end{center}
\end{table}

\begin{figure*}
\begin{centering}
\includegraphics[width=0.95\textwidth]{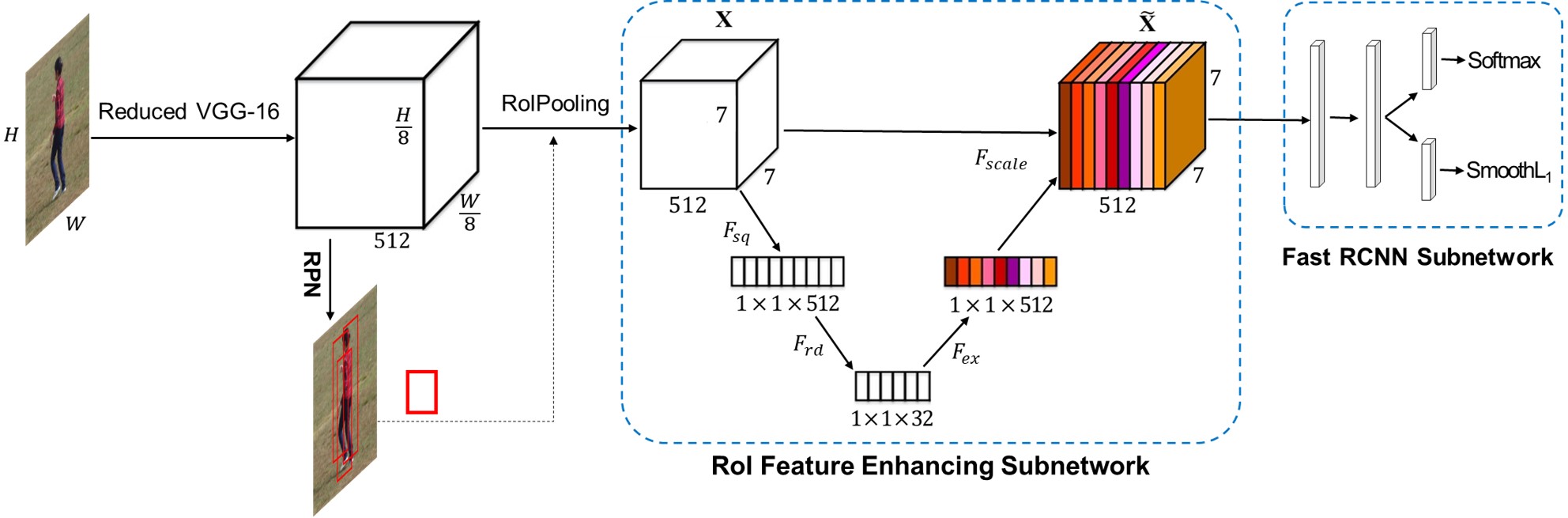}
\par\end{centering}
\caption{Diagram of our improved Faster R-CNN. Reduced VGG-16 means removing the fourth max pooling and using the ``hole algorithm''. RoI feature enhancing subnetwork is a reimplementation of SENet with an identical block structure, which consists of one global average pooling layer ($F_{sq}$) and two consecutive fully connected layers ($F_{rd}$ and $F_{ex}$). RoI feature {$\emph{X}$} is re-weighted to generate the output \emph{$\widetilde{X}$} of the SE block which then be fed directly into subsequent Fast R-CNN subnetwork.}
\label{fig:ifrcnn}
\end{figure*}

{\flushleft \textbf{Occlusion. }}
Occlusion is another important factor for evaluating the pedestrian detection performance. There are two types of occlusion: inter-class occlusion, which occurs when a person is occluded by stuff or objects from other categories; and intra-class occlusion (also referred to as crowd occlusion), which occurs when a person is occluded by other persons. As described in~\cite{DBLP:journals/corr/abs-1711-07752}, the intra-class occlusion is a more challenging issue than the inter-class occlusion. Lots of works have focused on the former problem and made great progress. However, the crowd occlusion has not been well researched and solved. On the one hand, it is difficult because of the problem itself. On the other hand, there is no suitable dataset. Therefore, we introduce this diversity and dense pedestrian detection dataset. To demonstrate its degree of crowd occlusion, we provide statistical information on pair-wise occlusion. For each image, we count the number of person pairs with different intersection over union (IoU) threshold. The results are shown in Table~\ref{tab:PairOverlap}. In average, few person pairs with an IoU threshold of $0.3$ are included in Caltech-USA. For CityPersons dataset, the number is less than one pair per image. However, the number is $9.21$ for WiderPerson. Moreover, there are averagely $2.15$ pairs whose IoU is greater than $0.5$ in the WiderPerson dataset. These data can demonstrate that various occlusion levels are well-represented in WiderPerson, especially heavily occluded cases, while they can be hardly found in previous datasets.

\subsection{Benchmarking}\label{subsec:benchmarking}
With the publication of this paper, we will create a website for WiderPerson dataset, where its annotations for the training and validation subsets are made freely available to academic and non-profit organizations for non-commercial, scientific use. There is also an evaluation instruction on the website for researchers to evaluate the performance of their detectors over the held-out testing annotations. A leaderboard will be maintained and results are tallied online, either by name or anonymous.

We follow the same evaluation metric as used for Caltech-USA~\cite{DBLP:journals/pami/DollarWSP12} and CityPersons~\cite{DBLP:conf/cvpr/ZhangBS17}, denoted as $MR$, which stands for the average log miss rate over false positives per-image ranging in $\left[ 10^{-2}, 10^0\right]$. $MR$ is a suitable indicator for the algorithms applied in the real world applications. When evaluating pedestrian detection performance, riders/partially-visible persons/crowd/ignore regions are ignored, which means that those annotations are not considered as false negatives and detections matching with those annotations are not counted as false positives.

\section{Provided Baseline Method}\label{4}
Before delving into our new dataset, we first build two strong baseline detectors as a tool for our experiment analyses based on Faster R-CNN~\cite{DBLP:journals/pami/RenHG017} and RetinaNet~\cite{DBLP:conf/iccv/LinPRK17}, which are two representative detectors from the two-stage and one-stage approach, respectively. We aim to find a straightforward architecture to provide good performance on WiderPerson.

\subsection{Improved Faster R-CNN}
Faster R-CNN is masterpieces of the detection framework for general object detection and has dominated the field of object detection in recent years. It essentially consists of two components: a fully convolutional Region Proposal Network (RPN) for proposing candidate regions which likely contain objects, followed by a downstream Fast R-CNN network to classify a region of image into objects (and background) and refine the boundaries of those regions. Although competitive performance has been achieved on general object detection task, it under-performs on the pedestrian detection task (as reported in~\cite{DBLP:conf/eccv/ZhangLLH16}). The reason behind its poor performance on pedestrian detection is that it fails to handle heavily occluded and dense pedestrians, which are dominant on our new dataset. In this work, we propose some improvements to extend the Faster RCNN architecture for occluded and dense pedestrian detection. Our improved Faster R-CNN is based on VGG-16~\cite{DBLP:journals/corr/SimonyanZ14a} and ResNet-50~\cite{DBLP:conf/cvpr/HeZRS16} as they are very common in pedestrian detection~\cite{DBLP:conf/eccv/ZhangLLH16, DBLP:conf/cvpr/ZhangBS17, DBLP:conf/cvpr/MaoXJC17}. We use the same anchor setting from~\cite{DBLP:conf/cvpr/ZhangBS17}, \ie, $11$ different anchor-box scales and $1$ aspect ratio ($w/h=0.41$) are used to capture objects across all sizes. Moreover, some improvements are proposed to boost the performance on pedestrian detection as follows. 

{\flushleft \textbf{Finer Feature Map.}}
The vanilla Faster R-CNN uses a coarse feature map as the detection layer, \ie, the last conv layer in the fifth block with stride of $16$ pixels. Having such a coarse stride is harmful to small pedestrian detection, since it reduces the chances of having a high score over pedestrian and forces the network to handle large displacement relative to the object appearance. To increase the feature map resolution, we remove the fourth down-sampling operation and reduce the stride from $16$ to $8$ pixels, helping the detector to handle small pedestrians. Specifically, all layers before the fourth down-sampling operation are unchanged and all convolutional filters after it are modified by the ``hole algorithm''~\cite{DBLP:conf/cvpr/LongSD15} (\ie, ``Algorithm $\grave{a}$ trous'') to compensate for the reduced stride.

{\flushleft \textbf{Ignore Region and Tiny Pedestrian Handling.}}
We implement an ignore region handling for Faster R-CNN. Ignore regions might contain objects of a given class without precise localization. Simply treating these regions as background introduces confusing samples and has a negative impact on the detector quality. The ignore region handling prevents the sampling of background boxes in those areas that could potentially overlap with real objects. Besides, training with very tiny samples could lead to models detecting a lot more false positives. Hence, we online filter pedestrians whose height is less than $20$ pixels after scaling during training. Filtered pedestrians are handled as ignore regions in order to ensure that they are not sampled as background during training.

{\flushleft \textbf{RoI Feature Enhancing.}}
The RoIPooling layer uses max pooling to convert the features inside any valid region of interest into a fixed-size feature map, which is used by subsequent Fast R-CNN network to further classify and regress the proposals for final detections. Therefore, the representational ability of the pooled feature is the key to achieve high performance, especially on our highly diverse dataset. Inspired by~\cite{DBLP:journals/corr/abs-1709-01507, DBLP:conf/cvpr/ZhangYS18}, we use a ``Squeeze-and-Excitation'' (SE) block to enhance the representational ability of the RoIPooling feature by explicitly modelling the interdependencies between the convolutional channels. More specific, the SE block performs sample-dependent feature re-weighting so as to select the more informative channel features while suppress less useful ones. As shown in Fig.~\ref{fig:ifrcnn}, the newly added SE block is composed of one global average pooling layer and two consecutive fully connected layers, which is easy to implement and can obtain remarkable improvements while add little additional computational costs.

{\flushleft \textbf{Dynamic Sample Strategy.}}
The vanilla Faster R-CNN has a fixed sample strategy, \ie, $256$ and $128$ samples for RPN and Fast R-CNN with $1:1$ and $1:3$ positive-negative ratio, respectively. Since there are $\sim\negmedspace 28.87$ persons per image in our dataset, the fixed sample strategy will lead to inadequate use of training positive samples. To solve this issue, we introduce a dynamic sample strategy: if there are too many positive samples, we determine the number of negative samples based on the above positive-negative ratio to ensure that all positive samples are used, otherwise we follow the original strategy.

\subsection{Vanilla RetinaNet}
In addition to the two-stage baseline detector, we also provide another baseline detector based on RetinaNet, the one-stage approach, which detects objects by regular and dense sampling over locations, scales and aspect ratios with high efficiency. RetinaNet proposes a focal loss to address the extreme foreground-background class imbalance by reshaping the standard cross entropy loss such that it down-weights the loss assigned to well-classified examples. We use the same setting of anchor scales as~\cite{DBLP:conf/iccv/LinPRK17} and only modify the height \emph{vs.} width ratio of anchors as $1$:$0.41$ in consideration of the pedestrian shape.

\section{Experiments} \label{5}
In this section, we will introduce our implementation details about data processing and training setting. Notably, all the experiments are conducted based on the improved Faster R-CNN with VGG-16 unless otherwise specified. Firstly, we verify the effectiveness of our improvements via model analysis. Then, we conduct some experiments to analyse our WiderPerson dataset in different aspects, including the detection result, quantity, quality and error. Finally, the generalization ability of our WiderPerson dataset will be evaluated on standard pedestrian benchmarks like Caltech-USA and CityPersons.

\subsection{Implementation Detail.}
{\flushleft \textbf{Data Processing. }}
To improve performance for small sized pedestrians, the input images are upscaled to a larger size using the bilinear interpolation algorithm. Specifically, the input image sizes of Caltech and CityPersons are set to $2\times$ and $1.3\times$ of the original images. As the images of WiderPerson are both collected from the Internet with various sizes, we resize the input so that their short edge is at $800$ pixels while the long edge should be no more than $1400$ pixels at the same time. We use horizontal image flipping as the only form of data augmentation. Multi-scale training and testing are not applied to ensure fair comparisons.

{\flushleft \textbf{Training Setting. }}
For the improved Faster R-CNN, all models are trained for $180k$ iterations with an initial learning rate of $0.01$, and decreased by a factor of $10$ after $120k$ on our WiderPerson dataset. On the CityPersons dataset, we set the learning rate to $10^{-3}$ for the first $40k$ iterations and decay it to $10^{-4}$ for another $20k$ iterations. On the Caltech-USA dataset, we train the network for $120k$ iterations with the initial learning rate $10^{-3}$ and decrease it by a factor of $10$ after the first $80k$ iterations. To fine-tune the improved Faster R-CNN from WiderPerson to Caltech-USA and CityPersons, the number of iterations is the same but the learning rate is halved overall. All these models are optimized by the Stochastic Gradient Descent (SGD) algorithm on $1$ TITAN X (Maxwell) GPU with a mini-batch $2$. Weight decay and momentum are set to $0.0005$ and $0.9$. Besides, the RetinaNet baseline on the WiderPerson dataset is trained with $16$ batch size for $25k$ iterations with $0.02$ initial learning rate, which is then divided by $10$ at $16k$ and again at $21k$ iterations. 

\begin{table}[b]
\centering
\caption{Analysis of proposed improvements. All models are based on Faster R-CNN with VGG-16, trained on WiderPerson {\tt training} set and tested on {\tt validation} set. Numbers indicate $MR$.}
\setlength{\tabcolsep}{4pt}
\begin{tabular}{c|cccccc}
\toprule[1.5pt]
\multicolumn{1}{c|}{Component} &\multicolumn{6}{c}{Step by step improvements} \\
\hline
new anchor setting & & \Checkmark & \Checkmark & \Checkmark & \Checkmark &\Checkmark \\
finer feature map & & & \Checkmark & \Checkmark & \Checkmark &\Checkmark \\
RoI feature enhancing & & & & \Checkmark & \Checkmark & \Checkmark\\
ignore region handling & & & & & \Checkmark & \Checkmark \\
dynamic sample strategy & & & & & & \Checkmark\\
\hline
{\em Easy} subset &43.01 &39.62 &35.12 & 31.45 & 29.98 &\textbf{29.61}\\
{\em Medium} subset &48.97 &46.28 &42.67 & 39.71 & 38.68 &\textbf{38.40}\\
{\em Hard} subset &55.62 &53.26 & 50.17 & 47.51 & 46.65 &\textbf{46.46}\\
\bottomrule[1.5pt]
\end{tabular}
\label{tab:ablation}
\end{table}

\begin{figure*}[t]
\centering
\includegraphics[width=0.98\textwidth]{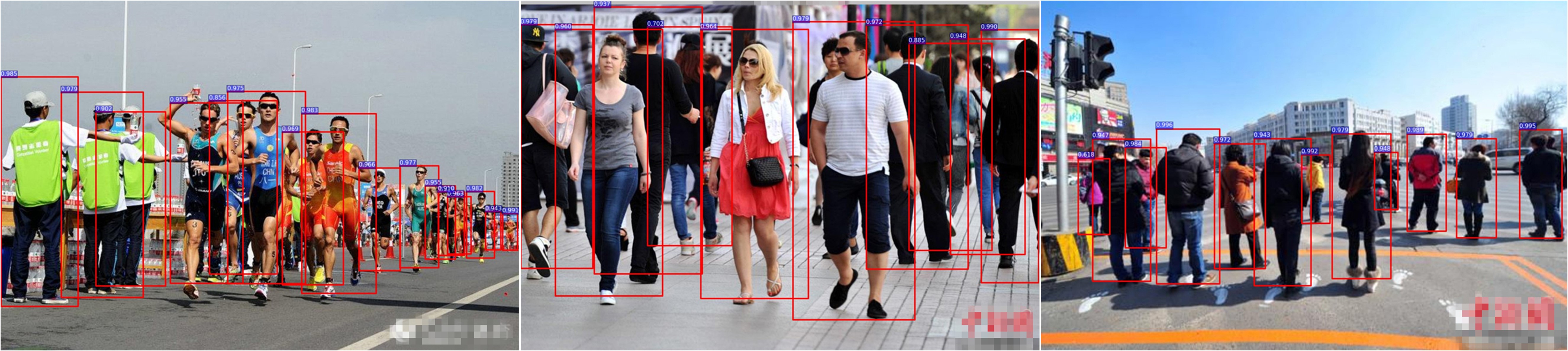}
\caption{Qualitative results for pedestrian detection of our improved Faster R-CNN with VGG-16 based on the WiderPerson dataset.}
\label{fig:result_frcnn}
\end{figure*}

\subsection{Model Analysis}
We carry out some ablation experiments on the WiderPerson validation subset to analyze our improved Faster R-CNN. For all the experiments, we use the same settings, except for specified changes to the components. We ablate each improvement one after another to examine how each proposed improvement affects the final performance. Firstly, we replace the dynamic sample strategy with the original strategy. Secondly, the RoI feature enhancing module is ablated. Thirdly, we do not handle ignore regions and tiny ground truths during training stage. Fourthly, we do not reduce the VGG-16 backbone. Finally, we use original anchor scales rather than new anchor setting~\cite{DBLP:conf/cvpr/ZhangBS17}.

Some promising conclusions can be summed up according to the ablative results in Tab.~\ref{tab:ablation}. Firstly, the new anchor setting is more suitable for the proposed dataset, which reduces the MR by $3.39\%$, $2.69\%$ and $2.36\%$ for Easy, Medium and Hard subset, respectively. Secondly, the finer feature map is used to provide more anchors and detailed information, which reduces the $MR$ $39.62\%$, $46.28\%$ and $53.26\%$ to $35.12\%$, $42.67\%$ and $50.17\%$ for Easy, Medium and Hard subset, respectively, demonstrating its effectiveness. Thirdly, the ignore region and tiny pedestrian handling is proposed to ignore small ground truths and prevent the sampling of background boxes in those ignored areas. The comparison between the second and third columns in Tab.~\ref{tab:ablation} demonstrates that it can bring $3.67\%$ (Easy), $2.96\%$ (Medium) and $2.66\%$ (Hard) drops in $MR$, attributing to not involving confusing samples in training. Fourthly, according to the third and fourth columns, we can observe a drop in $MR$ of $1.47\%$ (Easy), $1.03\%$ (Medium) and $0.86\%$ (Hard), these sharp declines demonstrate the effectiveness of the RoI feature enhancing. Finally, the comparison between the fourth and fifth columns in Tab.~\ref{tab:ablation} indicates that the dynamic sample strategy decreases the $MR$ by $0.37\%$ (Easy), $0.28\%$ (Medium) and $0.19\%$ (Hard), owning to making full use of training samples.

\subsection{Dataset Analysis}\label{sec:exp_ana}
All experiments in this subsection are trained based on WiderPerson training subset and the results are evaluated on the validation subset. Firstly, we evaluate our improved Faster R-CNN in detail on validation subset, then study on the quantity and quality, finally analyze common failure cases.

{\flushleft \textbf{Detection Results. }}
Table~\ref{tab:eval-frcnn} illustrates our baselines' results on the WiderPerson validation subset. On the one hand, we achieve promising $MR$ performances, \ie, $31.47\%$, $40.45\%$, $48.32\%$ for the vanilla RetinaNet, $29.61\%$, $38.40\%$, $46.46\%$ for the improved Faster R-CNN with VGG-16, and $28.75\%$, $37.82\%$, $46.06\%$ for the improved Faster R-CNN with ResNet-50 on Easy, Medium and Hard subsets, respectively. On the other hand, from these results, we can find that the proposed WiderPerson dataset is a challenging benchmark even for the state-of-the-art pedestrian detection algorithms. In Table~\ref{tab:caltech-usa} and Table~\ref{tab:citypersons}, we also report detection results of the improved Faster R-CNN with VGG-16 on Caltech, i.e., $5.49\%$ $MR$, and CityPersons, i.e., $12.49\%$ $MR$. It further demonstrates that our WiderPerson dataset is much challenging than the standard pedestrian detection benchmarks based on the detection performance. Since our WiderPerson dataset varies largely in scenario and occlusion, which bring many difficulties to pedestrian detection. The illustrative examples of pedestrian detection based on our improved Faster R-CNN with VGG-16 are shown in Fig.~\ref{fig:result_frcnn}.

\begin{table}[t]
\centering
\caption{$MR$ and speed performance of our baselines on the WiderPerson {\tt validation} subset.}
\label{tab:eval-frcnn}
\setlength{\tabcolsep}{6.5pt}
\begin{tabular}{cccccc}
\toprule[2pt]
Baseline &Backbone &FPS & Easy & Medium & Hard \\
\midrule
Vanilla RetinaNet & ResNet-50 &8.93 & {31.47} & {40.45} & {48.32} \\
Improved FRCNN & VGG-16 &0.83 & {29.61} & {38.40} & {46.46} \\
Improved FRCNN & ResNet-50 &0.77 & {28.75} & {37.82} & {46.06} \\
\bottomrule[2pt]
\end{tabular}
\end{table}

\begin{figure}[b]
\centering
\includegraphics[width=0.475\textwidth]{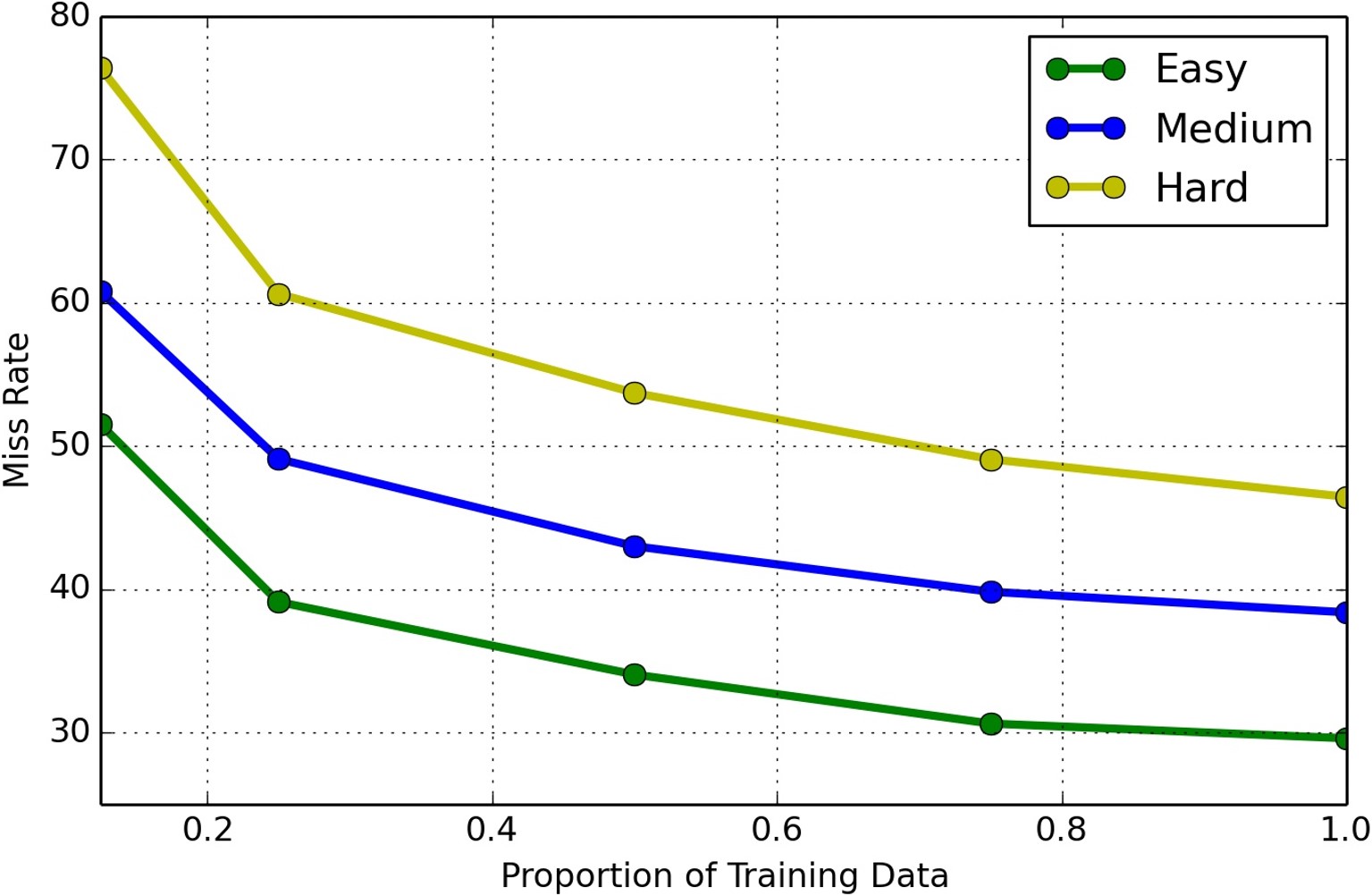}
\caption{Detection performance ($MR$) of our improved Faster R-CNN with VGG-16 as a function of training set size}
\label{fig:quantity}
\end{figure} 

{\flushleft \textbf{Quantity Analysis. }}
As indicated in~\cite{DBLP:conf/iccv/SunSSG17}, there is a logarithmic relation between the amount of training data and the performance of deep learning methods. To understand the impact of having a larger amount of training data, we show how the performance grows as training data increases on our benchmark. For this purpose, we train our baseline methods on different sized subsets which are randomly sampled from the training set. From Fig.\ref{fig:quantity} we can observe that logarithmic relation between training set size and detection performance also holds on our benchmark for the improved Faster R-CNN across three subsets, \ie, performance keeps improving with more data. Therefore, it is of great importance to provide CNNs with a large amount of data.

{\flushleft \textbf{Quality Analysis. }}
The importance of fine-grained annotations for riders and additional annotations for ignore regions is now examined. In ablation experiments, we have verified the effectiveness of ignored regions via training the mdoel without ignore region handling, it is in accordance with earlier findings~\cite{DBLP:conf/cvpr/ZhangBS17} that detection performance deteriorates when not using ignore regions during training. Besides, the evaluation protocol described in Section~\ref{subsec:benchmarking} ignores detected neighboring classes. For pedestrians this means that riders are not considered as false positives, hence training pedestrian detectors generally treat riders as ignore region. To verify whether rider annotations are useful for pedestrian detection, we can directly train the baseline detection method by including riders into pedestrians, since pedestrian and rider are annotated in the same way with a fixed aspect ratio on our dataset. As expected, comparing with training only with pedestrians, after these neighboring annotations are involved during training, detection performance increases on $MR$ from $29.61\%$, $38.40\%$ and $46.46\%$ to $29.35\%$, $38.27\%$ and $46.39\%$ for Easy, Medium and Hard subset, respectively. Therefore, adding fine-grained annotations for riders is helpful for the pedestrian detection performance, since we can treat them as an additional training samples.

\begin{figure}[t]
\centering
\includegraphics[width=0.35\textwidth]{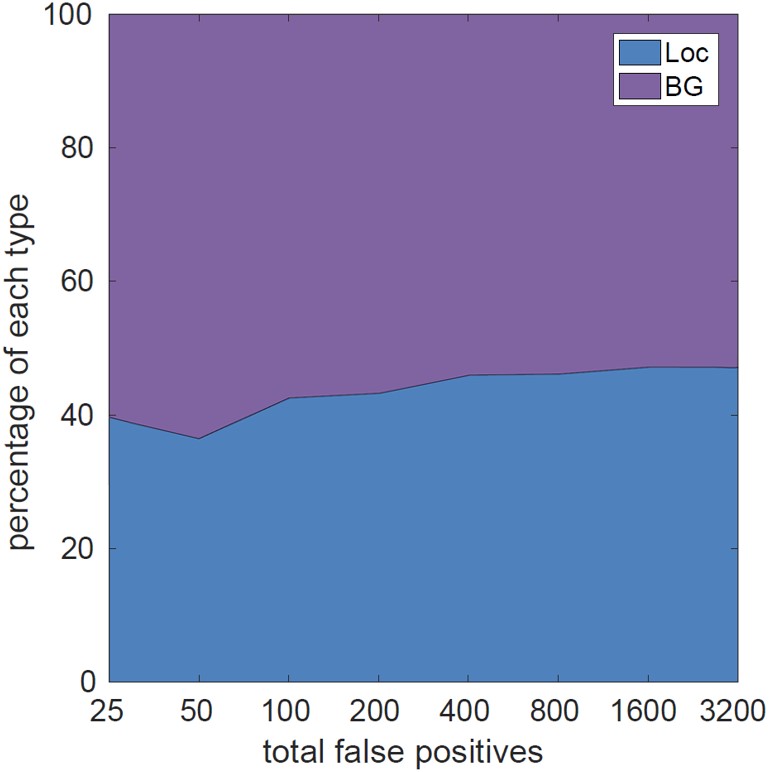}
\caption{Distribution of two error modes of false positives on the WiderPerson validation Hard subset.}
\label{fig:err_dist}
\end{figure}

\begin{figure*}[t]
\centering
\subfigure[False positive: a background region is mistakenly detected as a pedestrian, or a pedestrian is detected with a misaligned bounding box.]{
\label{fig:qualitative_mistakes_fp}
\includegraphics[width=0.957\textwidth]{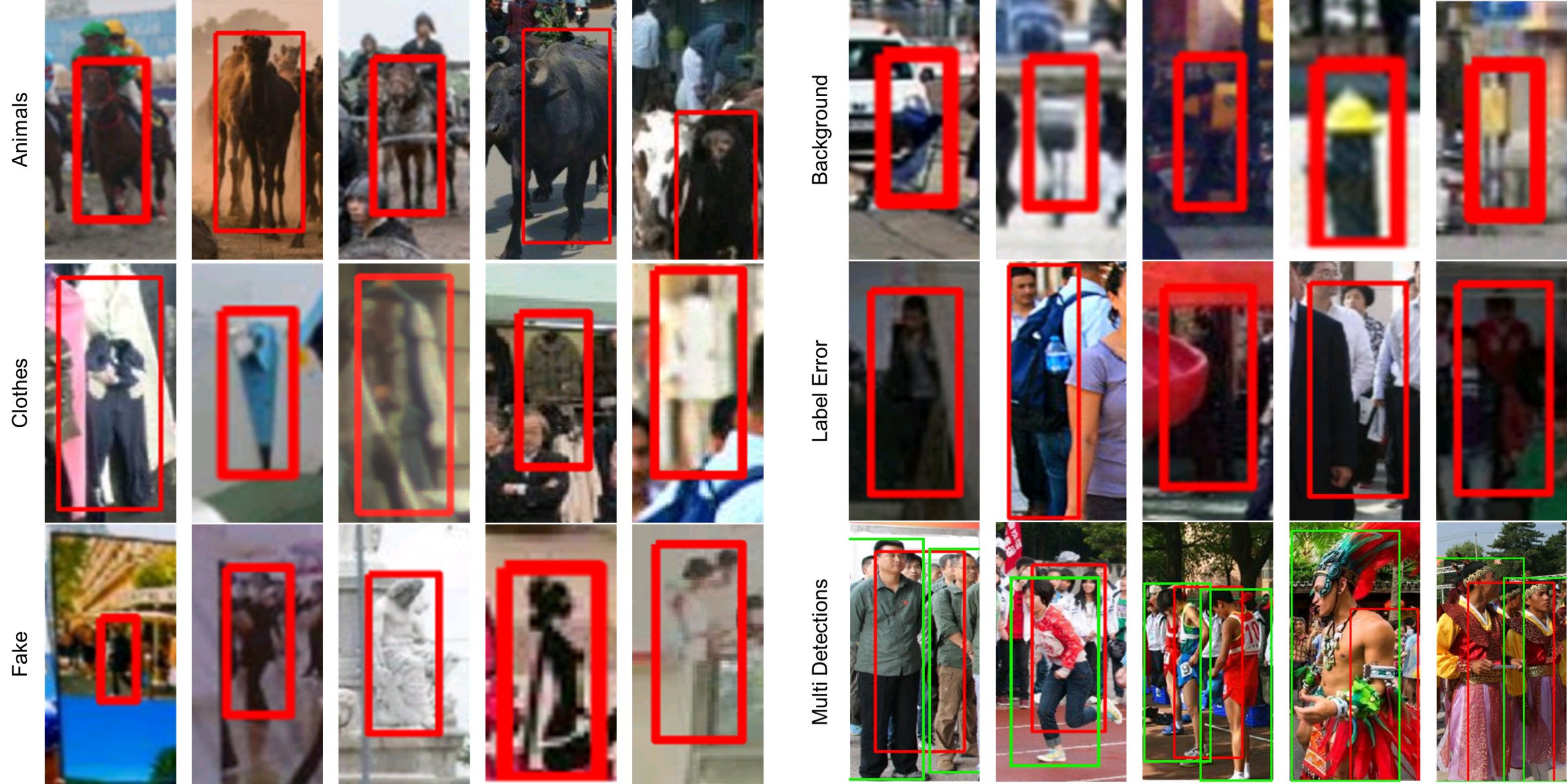}}
\subfigure[False negative: a pedestrian fails to be detected.]{
\label{fig:qualitative_mistakes_fn}
\includegraphics[width=0.957\textwidth]{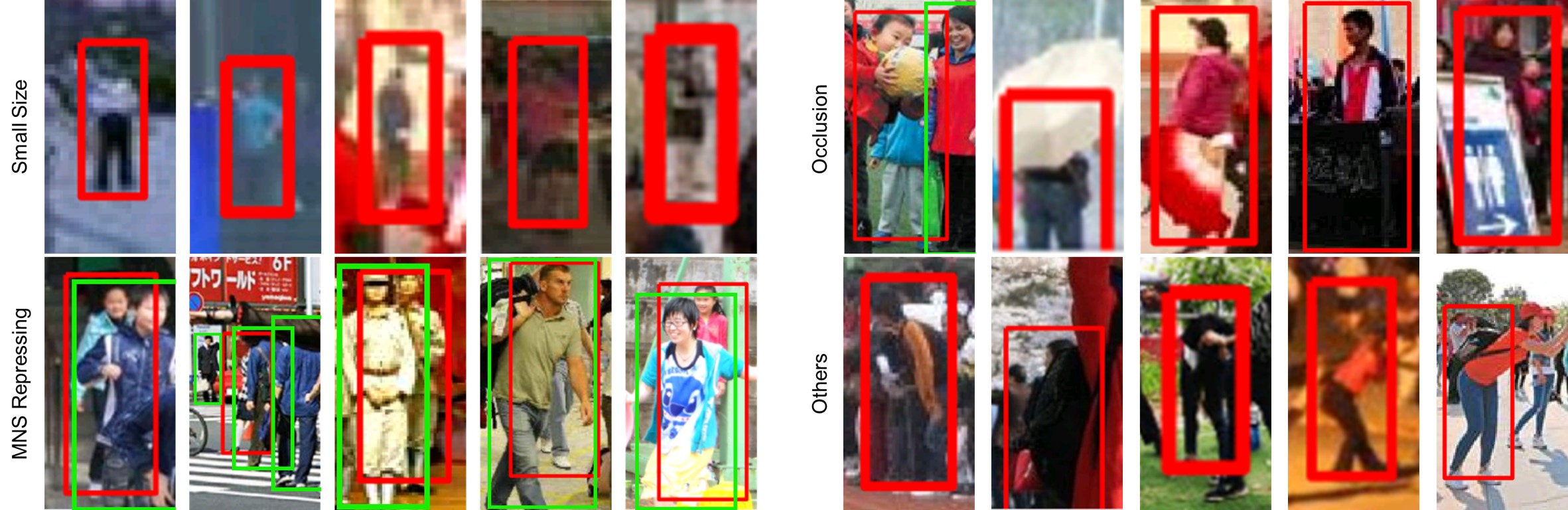}}
\caption{Qualitative detection errors for our improved Faster R-CNN (green: true positives, red: false positives or false negatives).}
\label{fig:qualitative_mistakes}
\end{figure*}

{\flushleft \textbf{Error Analysis. }}
We now utilize the detection analysis tool\footnote{\url{http://web.engr.illinois.edu/~dhoiem/projects/detectionAnalysis}} to analyze the detection errors of our improved Faster R-CNN qualitatively on our WiderPerson validation dataset. The detection errors consist of false positives and false negatives. Firstly, we analyse the false positive errors. There are two error modes of false positives in pedestrian detectors, \ie, location (LOC) and background (BG). LOC indicates the localization errors that occurs when a pedestrian is detected with a misaligned bounding box, and BG indicates that a background region is mistakenly detected as a pedestrian. Fig.~\ref{fig:err_dist} shows the distribution of two types of false positives and BG seems the dominating error mode among top-scoring detection. Figure~\ref{fig:qualitative_mistakes_fp} illustrates some qualitative false positives of this method. As can be seen, animals, clothes and fakes are principal sources for confusion with real pedestrians. Certain pedestrian poses and aspect ratios can lead to multiple detections for the same pedestrian as shown in the \textit{Multi Detections} category. Non-maximum suppression (NMS) is used by detection methods to suppress multiple detections. We use an \textit{IoU} threshold of $0.5$ which is not sufficient to suppress detections that have very diverse aspects.

Figure~\ref{fig:qualitative_mistakes_fn} illustrates some qualitative false negatives of this method. A lower \textit{IoU} threshold would lead to more false negatives. These already occur for an \textit{IoU} threshold of $0.5$ as shown in the \textit{NMS Repressing} category. Because of the high \textit{IoU} between pedestrians, not all of them can be detected because of the greedy NMS. Thus, NMS is an important part of many deep learning methods that is usually not trained but has a great influence on detection performance. Small and occluded pedestrians are a further common source for false negatives. These two groups have also been analyzed in~\cite{DBLP:conf/itsc/RajaramOT15}. In some scenarios, usually only the lower part of a pedestrian is occluded due to various obstacles. In our qualitative analysis we have false negatives where the head is occluded. These are particularly challenging for pedestrian detection methods, as these cases are quite rare in the training dataset. Further challenges are unusual and extreme poses as shown in the \textit{Others} group.

\subsection{Generalization Capability}
In this subsection, we evaluate the generalization capability of our WiderPerson dataset. As illustrated in Section~\ref{sec:dataset_stat}, the size of WiderPerson dataset is obviously more diverse and larger than the existing benchmarks, like Caltech-USA~\cite{DBLP:journals/pami/DollarWSP12} and CityPersons~\cite{DBLP:conf/cvpr/ZhangBS17}. Naturally, our dataset, with a reduced bias, should better capture the true world and result in superior generalization capabilities of the detectors which are trained on this dataset. To demonstrate the increased diversity of our dataset, we first train the model on our WiderPerson dataset and then fine-tune it on the other pedestrian detection benchmarks.

{\flushleft \textbf{Caltech. }}
The Caltech-USA dataset is one of the most popular and challenging datasets for pedestrian detection, which comes from approximately $10$ hours $30$Hz VGA video recorded by a car traversing the streets in the greater Los Angeles metropolitan area. We use the new high quality annotations provided by~\cite{DBLP:conf/cvpr/ZhangBOHS16} to train and evaluate. The training and testing sets contains $42,782$ and $4,024$ frames, respectively. The results are shown for the Caltech-USA dataset in Table \ref{tab:caltech-usa}. The overall detection performance is superior for the cases in which WiderPerson is used for pre-training. Our improved Faster R-CNN achieves $5.49\%$ $MR$ for pedestrians on the Caltech-USA testing set for the reasonable setting. When we directly evaluate the Caltech-USA trained model on the proposed WiderPerson Easy subset, we get a very high MR of $82.79\%$ since Caltech-USA has limited density and diversity. In contrast, our model trained on WiderPerson without fine-tuning achieves $9.72\%$ $MR$ and can be boost to $4.27\%$ $MR$ with fine-tuning. Based on the pre-training of our WiderPerson dataset, our algorithm has superior performance on the Caltech-USA benchmark against the one without WiderPerson pre-training, and performs on-pair with the state-of-the-arts.

\begin{table}[t]
\begin{center}
\caption{Experimental results on Caltech-USA.}
\label{tab:caltech-usa}
\setlength{\tabcolsep}{13pt}
\begin{tabular}{ccc}
\toprule[2pt]
{Training} & {Testing} & $MR$ \\
\hline
Caltech-USA & Caltech-USA & {5.49}\\
Caltech-USA & WiderPerson (Easy) & {82.79}\\
WiderPerson & Caltech-USA & {9.72}\\
WiderPerson$\Rightarrow$Caltech-USA & Caltech-USA & {4.27}\\
\hline
\hline
\multicolumn{2}{c}{RPN$+$BF~\cite{DBLP:conf/eccv/ZhangLLH16}} &7.3\\
\multicolumn{2}{c}{HyperLearner~\cite{DBLP:conf/cvpr/MaoXJC17}} &5.5\\
\multicolumn{2}{c}{OR-RCNN~\cite{zhang2018occlusion}} &4.1\\
\multicolumn{2}{c}{Repulsion Loss~\cite{DBLP:journals/corr/abs-1711-07752}} &4.0\\
\bottomrule[2pt]
\end{tabular}
\end{center}
\end{table}

{\flushleft \textbf{CityPersons. }}
The CityPersons dataset is built upon the semantic segmentation dataset Cityscapes to provide a new dataset of interest for pedestrian detection. It is recorded across $18$ different cities in Germany with $3$ different seasons and various weather conditions. The dataset includes $5,000$ images ($2,975$ for training, $500$ for validation, and $1,525$ for testing) with $35,000$ manually annotated persons plus $\sim\negmedspace 13,000$ ignore region annotations. Both the bounding boxes and visible parts of pedestrians are provided and there are approximately $7$ pedestrians in average per image. The results are shown for the CityPersons dataset in Table \ref{tab:citypersons}. The same findings hold for the CityPersons benchmark. Training on CityPersons dataset and testing on WiderPerson Easy subset has $73.45\%$ $MR$, while training on WiderPerson dataset and testing on CityPersons validation subset achieves $16.17\%$ $MR$, indicating the difficulty and expandability of our dataset. Again, our improved Faster R-CNN model pre-trained on WiderPerson can reduce the $MR$ from $12.49\%$ to $11.13\%$ that is on-pair with the state-of-the-arts, demonstrating our WiderPerson dataset can serve as an effective pre-training dataset for pedestrian detection task.

\begin{table}[t]
\begin{center}
\caption{Experimental results on CityPersons.}
\label{tab:citypersons}
\setlength{\tabcolsep}{12.5pt}
\begin{tabular}{ccc}
\toprule[2pt]
{Training} & {Testing} & $MR$ \\
\hline
CityPersons & CityPersons & {12.49}\\
CityPersons & WiderPerson (Easy) & {73.45}\\
WiderPerson & CityPersons & {16.17}\\
WiderPerson$\Rightarrow$CityPersons & CityPersons & 11.13\\
\hline
\hline
\multicolumn{2}{c}{Adapted Faster RCNN~\cite{DBLP:conf/cvpr/ZhangBS17}} &12.8\\
\multicolumn{2}{c}{Repulsion Loss~\cite{DBLP:journals/corr/abs-1711-07752}} &11.6\\
\multicolumn{2}{c}{OR-CNN\cite{zhang2018occlusion}} &11.0\\
\bottomrule[2pt]
\end{tabular}
\end{center}
\end{table}

{\flushleft \textbf{Summary. }}
The superior performance on both Caltech-USA and CityPersons datasets when using WiderPerson for pre-training indicates a high dataset diversity. Models trained on this dataset will have increased generalization capabilities. However, due to the dataset biases, solely training on a dataset from the other domain without fine-tuning results in worse detection performance. Despite the dataset biases, the models are able to learn general features for the task of pedestrians detection when pre-trained on WiderPerson which proves useful for other datasets as well after fine-tuning. Using transfer learning to pre-train a network on generic data and fine-tune on the target domain is widely applied and used to increase overall performance.

\section{Conclusion} \label{6}
Current pedestrian detection benchmark datasets have contributed to spurring interest and progress in pedestrian detection research. With the help of CNN, modern methods have achieved remarkable performance on these benchmarks. However, it is still difficult to assess for real world performance, since there is a gap in the diversity and density between existing pedestrian detection benchmarks and real world requirements: 1) most of current datasets are collected in the fixed traffic scenario, which significantly reduces the diversity of the foreground and background. 2) crowd scenarios with occluded pedestrian are still under represented, limiting the variations in density. These limitations have partially contributed to the failure of some algorithms in coping with heavy occlusion and atypical scenario. To move forward the field of pedestrian detection, we introduce a diverse and dense pedestrian detection dataset called WiderPerson, which consists of $13,382$ images with $399,786$ annotations and varies largely in scenario and occlusion. Providing high quality annotations, it enables new experiments both for training better models and as new test benchmark. We propose some strong baseline detectors based on Faster R-CNN and RetinaNet to benchmark the state-of-the-art detector. The cross-dataset generalization results of WiderPerson dataset demonstrate that it is an effective training source for pedestrian detection and can help to achieve state-of-the-art performance on the Caltech-USA and CityPersons datasets. In the future, we will provide continuous improvements and additions to the WiderPerson dataset. Besides, we plan to annotate the head bounding box for each pedestrian and explore their relationship to facilitate further studies on the dense pedestrian detection.

\section*{Acknowledgment}
This work was supported by the National Key Research and Development Plan (Grant No.2016YFC0801002), the Chinese National Natural Science Foundation Projects $\#$61876179, $\#$61872367, $\#$61806203, Science and Technology Development Fund of Macau (No. 152/2017/A, 0025/2018/A1, 008/2019/A1). We also acknowledge the support of NVIDIA with the GPU donation for this research.

\ifCLASSOPTIONcaptionsoff
  \newpage
\fi

\bibliographystyle{IEEEtran}
\bibliography{IEEEabrv,reference}

\begin{thebibliography}{10}
\providecommand{\url}[1]{#1}
\csname url@samestyle\endcsname
\providecommand{\newblock}{\relax}
\providecommand{\bibinfo}[2]{#2}
\providecommand{\BIBentrySTDinterwordspacing}{\spaceskip=0pt\relax}
\providecommand{\BIBentryALTinterwordstretchfactor}{4}
\providecommand{\BIBentryALTinterwordspacing}{\spaceskip=\fontdimen2\font plus
\BIBentryALTinterwordstretchfactor\fontdimen3\font minus
  \fontdimen4\font\relax}
\providecommand{\BIBforeignlanguage}[2]{{%
\expandafter\ifx\csname l@#1\endcsname\relax
\typeout{** WARNING: IEEEtran.bst: No hyphenation pattern has been}%
\typeout{** loaded for the language `#1'. Using the pattern for}%
\typeout{** the default language instead.}%
\else
\language=\csname l@#1\endcsname
\fi
#2}}
\providecommand{\BIBdecl}{\relax}
\BIBdecl

\bibitem{DBLP:journals/pami/DollarWSP12}
P.~Doll{\'{a}}r, C.~Wojek, B.~Schiele, and P.~Perona, ``Pedestrian detection:
  An evaluation of the state of the art,'' \emph{TPAMI}, pp. 743--761, 2012.

\bibitem{DBLP:conf/cvpr/GeigerLU12}
A.~Geiger, P.~Lenz, and R.~Urtasun, ``Are we ready for autonomous driving? the
  {KITTI} vision benchmark suite,'' in \emph{CVPR}, 2012, pp. 3354--3361.

\bibitem{DBLP:conf/cvpr/ZhangBS17}
S.~Zhang, R.~Benenson, and B.~Schiele, ``Citypersons: {A} diverse dataset for
  pedestrian detection,'' in \emph{CVPR}, 2017, pp. 4457--4465.

\bibitem{DBLP:conf/cvpr/ZhangYS18}
S.~Zhang, J.~Yang, and B.~Schiele, ``Occluded pedestrian detection through
  guided attention in cnns,'' in \emph{CVPR}, 2018.

\bibitem{DBLP:journals/corr/abs-1711-07752}
X.~Wang, T.~Xiao, Y.~Jiang, S.~Shao, J.~Sun, and C.~Shen, ``Repulsion loss:
  Detecting pedestrians in a crowd,'' in \emph{CVPR}, 2018.

\bibitem{wang2018pedestrian}
S.~Wang, J.~Cheng, H.~Liu, F.~Wang, and H.~Zhou, ``Pedestrian detection via
  body-part semantic and contextual information with dnn,'' \emph{TMM}, 2018.

\bibitem{DBLP:journals/tmm/LiLSXFY18}
J.~Li, X.~Liang, S.~Shen, T.~Xu, J.~Feng, and S.~Yan, ``Scale-aware fast
  {R-CNN} for pedestrian detection,'' \emph{TMM}, vol.~20, no.~4, pp. 985--996,
  2018.

\bibitem{zhang2018occlusion}
S.~Zhang, L.~Wen, X.~Bian, Z.~Lei, and S.~Z. Li, ``Occlusion-aware r-cnn:
  Detecting pedestrians in a crowd,'' in \emph{ECCV}, 2018.

\bibitem{DBLP:conf/cvpr/IdreesSSS13}
H.~Idrees, I.~Saleemi, C.~Seibert, and M.~Shah, ``Multi-source multi-scale
  counting in extremely dense crowd images,'' in \emph{CVPR}, 2013, pp.
  2547--2554.

\bibitem{DBLP:conf/cvpr/ShaoKLW15}
J.~Shao, K.~Kang, C.~C. Loy, and X.~Wang, ``Deeply learned attributes for
  crowded scene understanding,'' in \emph{CVPR}, 2015, pp. 4657--4666.

\bibitem{DBLP:journals/pami/RenHG017}
S.~Ren, K.~He, R.~B. Girshick, and J.~Sun, ``Faster {R-CNN:} towards real-time
  object detection with region proposal networks,'' \emph{TPAMI}, vol.~39,
  no.~6, pp. 1137--1149, 2017.

\bibitem{DBLP:conf/ivs/SilbersteinLKG14}
S.~Silberstein, D.~Levi, V.~Kogan, and R.~Gazit, ``Vision-based pedestrian
  detection for rear-view cameras,'' in \emph{Intelligent Vehicles Symposium},
  2014, pp. 853--860.

\bibitem{DBLP:conf/cvpr/DalalT05}
N.~Dalal and B.~Triggs, ``Histograms of oriented gradients for human
  detection,'' in \emph{CVPR}, 2005, pp. 886--893.

\bibitem{DBLP:conf/iccv/WuN07}
B.~Wu and R.~Nevatia, ``Cluster boosted tree classifier for multi-view,
  multi-pose object detection,'' in \emph{ICCV}, 2007, pp. 1--8.

\bibitem{DBLP:conf/iccv/EssLG07}
A.~Ess, B.~Leibe, and L.~J.~V. Gool, ``Depth and appearance for mobile scene
  analysis,'' in \emph{ICCV}, 2007, pp. 1--8.

\bibitem{geronimo2007adaptive}
D.~Ger{\'o}nimo, A.~Sappa, A.~L{\'o}pez, and D.~Ponsa, ``Adaptive image
  sampling and windows classification for on-board pedestrian detection,'' in
  \emph{ICVS}, vol.~39, 2007.

\bibitem{overett2008new}
G.~Overett, L.~Petersson, N.~Brewer, L.~Andersson, and N.~Pettersson, ``A new
  pedestrian dataset for supervised learning,'' in \emph{Intelligent Vehicles
  Symposium}, 2008, pp. 373--378.

\bibitem{DBLP:journals/pami/EnzweilerG09}
M.~Enzweiler and D.~M. Gavrila, ``Monocular pedestrian detection: Survey and
  experiments,'' \emph{TPAMI}, vol.~31, no.~12, pp. 2179--2195, 2009.

\bibitem{DBLP:conf/cvpr/WojekWS09}
C.~Wojek, S.~Walk, and B.~Schiele, ``Multi-cue onboard pedestrian detection,''
  in \emph{Computer Society Workshop on CVPR}, 2009, pp. 794--801.

\bibitem{DBLP:conf/ivs/LiFYXBPLG16}
X.~Li, F.~Flohr, Y.~Yang, H.~Xiong, M.~Braun, S.~Pan, K.~Li, and D.~M. Gavrila,
  ``A new benchmark for vision-based cyclist detection,'' in \emph{Intelligent
  Vehicles Symposium (IV)}, 2016, pp. 1028--1033.

\bibitem{DBLP:conf/cvpr/HwangPKCK15}
S.~Hwang, J.~Park, N.~Kim, Y.~Choi, and I.~S. Kweon, ``Multispectral pedestrian
  detection: Benchmark dataset and baseline,'' in \emph{CVPR}, 2015, pp.
  1037--1045.

\bibitem{DBLP:journals/pami/ZhangBOHS18}
S.~Zhang, R.~Benenson, M.~Omran, J.~H. Hosang, and B.~Schiele, ``Towards
  reaching human performance in pedestrian detection,'' \emph{TPAMI}, vol.~40,
  no.~4, pp. 973--986, 2018.

\bibitem{DBLP:journals/corr/abs-1805-07193}
M.~Braun, S.~Krebs, F.~Flohr, and D.~M. Gavrila, ``The eurocity persons
  dataset: {A} novel benchmark for object detection,'' \emph{CoRR}, 2018.

\bibitem{DBLP:journals/tip/SundaresanC09}
A.~Sundaresan and R.~Chellappa, ``Multicamera tracking of articulated human
  motion using shape and motion cues,'' \emph{TIP}, vol.~18, no.~9, pp.
  2114--2126, 2009.

\bibitem{DBLP:journals/tip/LanZYC18}
X.~Lan, S.~Zhang, P.~C. Yuen, and R.~Chellappa, ``Learning common and
  feature-specific patterns: {A} novel multiple-sparse-representation-based
  tracker,'' \emph{TIP}, vol.~27, no.~4, pp. 2022--2037, 2018.

\bibitem{DBLP:conf/eccv/LiuAESRFB16}
W.~Liu, D.~Anguelov, D.~Erhan, C.~Szegedy, S.~E. Reed, C.~Fu, and A.~C. Berg,
  ``{SSD:} single shot multibox detector,'' in \emph{ECCV}, 2016, pp. 21--37.

\bibitem{DBLP:conf/cvpr/ZhangSF18}
S.~Zhang, L.~Wen, X.~Bian, Z.~Lei, and S.~Z. Li, ``Single-shot refinement
  neural network for object detection,'' in \emph{CVPR}, 2018.

\bibitem{DBLP:journals/tmm/LiWLDXFY17}
J.~Li, Y.~Wei, X.~Liang, J.~Dong, T.~Xu, J.~Feng, and S.~Yan, ``Attentive
  contexts for object detection,'' \emph{TMM}, vol.~19, no.~5, pp. 944--954,
  2017.

\bibitem{DBLP:conf/bmvc/DollarTPB09}
P.~Doll{\'{a}}r, Z.~Tu, P.~Perona, and S.~J. Belongie, ``Integral channel
  features,'' in \emph{BMVC}, 2009, pp. 1--11.

\bibitem{DBLP:conf/cvpr/YanZLLL13}
J.~Yan, X.~Zhang, Z.~Lei, S.~Liao, and S.~Z. Li, ``Robust multi-resolution
  pedestrian detection in traffic scenes,'' in \emph{CVPR}, 2013, pp.
  3033--3040.

\bibitem{DBLP:conf/cvpr/ZhangBC14}
S.~Zhang, C.~Bauckhage, and A.~B. Cremers, ``Informed haar-like features
  improve pedestrian detection,'' in \emph{CVPR}, 2014, pp. 947--954.

\bibitem{DBLP:journals/pami/DollarABP14}
P.~Doll{\'{a}}r, R.~Appel, S.~J. Belongie, and P.~Perona, ``Fast feature
  pyramids for object detection,'' \emph{TPAMI}, vol.~36, no.~8, pp.
  1532--1545, 2014.

\bibitem{DBLP:conf/cvpr/ZhangBS15}
S.~Zhang, R.~Benenson, and B.~Schiele, ``Filtered channel features for
  pedestrian detection,'' in \emph{CVPR}, 2015, pp. 1751--1760.

\bibitem{DBLP:conf/eccv/PaisitkriangkraiSH14}
S.~Paisitkriangkrai, C.~Shen, and A.~van~den Hengel, ``Strengthening the
  effectiveness of pedestrian detection with spatially pooled features,'' in
  \emph{ECCV}, 2014, pp. 546--561.

\bibitem{DBLP:conf/cvpr/SermanetKCL13}
P.~Sermanet, K.~Kavukcuoglu, S.~Chintala, and Y.~LeCun, ``Pedestrian detection
  with unsupervised multi-stage feature learning,'' in \emph{CVPR}, 2013, pp.
  3626--3633.

\bibitem{DBLP:conf/cvpr/HosangOBS15}
J.~H. Hosang, M.~Omran, R.~Benenson, and B.~Schiele, ``Taking a deeper look at
  pedestrians,'' in \emph{CVPR}, 2015, pp. 4073--4082.

\bibitem{DBLP:conf/cvpr/TianLWT15}
Y.~Tian, P.~Luo, X.~Wang, and X.~Tang, ``Pedestrian detection aided by deep
  learning semantic tasks,'' in \emph{CVPR}, 2015, pp. 5079--5087.

\bibitem{DBLP:conf/iccv/BrazilYL17}
G.~Brazil, X.~Yin, and X.~Liu, ``Illuminating pedestrians via simultaneous
  detection and segmentation,'' in \emph{ICCV}, 2017, pp. 4960--4969.

\bibitem{DBLP:conf/iccv/CaiSV15}
Z.~Cai, M.~J. Saberian, and N.~Vasconcelos, ``Learning complexity-aware
  cascades for deep pedestrian detection,'' in \emph{ICCV}, 2015, pp.
  3361--3369.

\bibitem{DBLP:conf/bmvc/AngelovaKVOF15}
A.~Angelova, A.~Krizhevsky, V.~Vanhoucke, A.~S. Ogale, and D.~Ferguson,
  ``Real-time pedestrian detection with deep network cascades,'' in
  \emph{BMVC}, 2015, pp. 32.1--32.12.

\bibitem{DBLP:conf/cvpr/YangCL16}
F.~Yang, W.~Choi, and Y.~Lin, ``Exploit all the layers: Fast and accurate {CNN}
  object detector with scale dependent pooling and cascaded rejection
  classifiers,'' in \emph{CVPR}, 2016.

\bibitem{DBLP:conf/eccv/ZhangLLH16}
L.~Zhang, L.~Lin, X.~Liang, and K.~He, ``Is faster {R-CNN} doing well for
  pedestrian detection?'' in \emph{ECCV}, 2016, pp. 443--457.

\bibitem{DBLP:conf/eccv/CaiFFV16}
Z.~Cai, Q.~Fan, R.~S. Feris, and N.~Vasconcelos, ``A unified multi-scale deep
  convolutional neural network for fast object detection,'' in \emph{ECCV},
  2016, pp. 354--370.

\bibitem{DBLP:conf/cvpr/MaoXJC17}
J.~Mao, T.~Xiao, Y.~Jiang, and Z.~Cao, ``What can help pedestrian detection?''
  in \emph{CVPR}, 2017, pp. 6034--6043.

\bibitem{DBLP:conf/cvpr/OuyangW12}
W.~Ouyang and X.~Wang, ``A discriminative deep model for pedestrian detection
  with occlusion handling,'' in \emph{CVPR}, 2012, pp. 3258--3265.

\bibitem{DBLP:conf/iccv/TianLWT15}
Y.~Tian, P.~Luo, X.~Wang, and X.~Tang, ``Deep learning strong parts for
  pedestrian detection,'' in \emph{ICCV}, 2015, pp. 1904--1912.

\bibitem{DBLP:conf/accv/ZhouY16}
C.~Zhou and J.~Yuan, ``Learning to integrate occlusion-specific detectors for
  heavily occluded pedestrian detection,'' in \emph{ACCV}, 2016, pp. 305--320.

\bibitem{DBLP:conf/iccv/WuN05}
B.~Wu and R.~Nevatia, ``Detection of multiple, partially occluded humans in a
  single image by bayesian combination of edgelet part detectors,'' in
  \emph{ICCV}, 2005, pp. 90--97.

\bibitem{DBLP:conf/eccv/DuanAL10}
G.~Duan, H.~Ai, and S.~Lao, ``A structural filter approach to human
  detection,'' in \emph{ECCV}, 2010, pp. 238--251.

\bibitem{DBLP:conf/cvpr/LeibeSS05}
B.~Leibe, E.~Seemann, and B.~Schiele, ``Pedestrian detection in crowded
  scenes,'' in \emph{CVPR}, 2005, pp. 878--885.

\bibitem{DBLP:conf/iccv/WangHY09}
X.~Wang, T.~X. Han, and S.~Yan, ``An {HOG-LBP} human detector with partial
  occlusion handling,'' in \emph{ICCV}, 2009, pp. 32--39.

\bibitem{DBLP:conf/cvpr/OuyangW13}
W.~Ouyang and X.~Wang, ``Single-pedestrian detection aided by multi-pedestrian
  detection,'' in \emph{CVPR}, 2013, pp. 3198--3205.

\bibitem{DBLP:conf/cvpr/PepikSGS13}
B.~Pepik, M.~Stark, P.~V. Gehler, and B.~Schiele, ``Occlusion patterns for
  object class detection,'' in \emph{CVPR}, 2013, pp. 3286--3293.

\bibitem{DBLP:conf/iccv/ZhouY17}
C.~Zhou and J.~Yuan, ``Multi-label learning of part detectors for heavily
  occluded pedestrian detection,'' in \emph{ICCV}, 2017, pp. 3506--3515.

\bibitem{DBLP:conf/ccs/MihcakV01}
M.~K. Mih{\c{c}}ak and R.~Venkatesan, ``New iterative geometric methods for
  robust perceptual image hashing,'' in \emph{Security and Privacy in Digital
  Rights Management, ACM Workshop}, 2001, pp. 13--21.

\bibitem{DBLP:conf/cvpr/YangLLT16}
S.~Yang, P.~Luo, C.~C. Loy, and X.~Tang, ``{WIDER} {FACE:} {A} face detection
  benchmark,'' in \emph{CVPR}, 2016, pp. 5525--5533.

\bibitem{DBLP:conf/eccv/ZitnickD14}
C.~L. Zitnick and P.~Doll{\'{a}}r, ``Edge boxes: Locating object proposals from
  edges,'' in \emph{ECCV}, 2014, pp. 391--405.

\bibitem{DBLP:conf/iccv/LinPRK17}
T.~Lin, P.~Goyal, R.~B. Girshick, K.~He, and P.~Doll{\'{a}}r, ``Focal loss for
  dense object detection,'' in \emph{ICCV}, 2017.

\bibitem{DBLP:journals/corr/SimonyanZ14a}
K.~Simonyan and A.~Zisserman, ``Very deep convolutional networks for
  large-scale image recognition,'' \emph{CoRR}, vol. abs/1409.1556, 2014.

\bibitem{DBLP:conf/cvpr/HeZRS16}
K.~He, X.~Zhang, S.~Ren, and J.~Sun, ``Deep residual learning for image
  recognition,'' in \emph{CVPR}, 2016, pp. 770--778.

\bibitem{DBLP:conf/cvpr/LongSD15}
J.~Long, E.~Shelhamer, and T.~Darrell, ``Fully convolutional networks for
  semantic segmentation,'' in \emph{CVPR}, 2015, pp. 3431--3440.

\bibitem{DBLP:journals/corr/abs-1709-01507}
J.~Hu, L.~Shen, and G.~Sun, ``Squeeze-and-excitation networks,'' in
  \emph{CVPR}, 2018.

\bibitem{DBLP:conf/iccv/SunSSG17}
C.~Sun, A.~Shrivastava, S.~Singh, and A.~Gupta, ``Revisiting unreasonable
  effectiveness of data in deep learning era,'' in \emph{ICCV}, 2017, pp.
  843--852.

\bibitem{DBLP:conf/itsc/RajaramOT15}
R.~N. Rajaram, E.~Ohn{-}Bar, and M.~M. Trivedi, ``An exploration of why and
  when pedestrian detection fails,'' in \emph{ITSC}, 2015, pp. 2335--2340.

\bibitem{DBLP:conf/cvpr/ZhangBOHS16}
S.~Zhang, R.~Benenson, M.~Omran, J.~H. Hosang, and B.~Schiele, ``How far are we
  from solving pedestrian detection?'' in \emph{CVPR}, 2016.

\end{thebibliography}

\end{document}